\documentclass{article} 
\usepackage{iclr2024_conference,times}

\usepackage{amsmath,amsfonts,bm}

\def\eqref#1{equation~\ref{#1}}

\def\1{\bm{1}}

\DeclareMathAlphabet{\mathsfit}{\encodingdefault}{\sfdefault}{m}{sl}
\SetMathAlphabet{\mathsfit}{bold}{\encodingdefault}{\sfdefault}{bx}{n}

\usepackage{hyperref}
\usepackage{url}
\usepackage{graphicx}
\usepackage{booktabs}
\usepackage{multirow}
\usepackage{subcaption}
\usepackage{makecell}
\usepackage{xcolor}

\newcommand{\Red}[1]{\textcolor[rgb]{1.00,0.00,0.00}{#1}}

\newcommand{\Green}[1]{\textcolor[rgb]{0.00,0.80,0.00}{#1}}

\newcommand{\update}[1]{\textcolor[rgb]{0.00,0.00,0.00}{#1}}

\newcommand{\tacticBlue}[1]{\textcolor[rgb]{0.0, 0.1, 0.6}{#1}}

\newcommand*\diff{\mathop{}\!\mathrm{d}}  

\newcommand{\ourmethod}[0]{\textsc{Mustard}}
\newcommand{\ourdataset}[0]{\textsc{MustardSauce}}
\newcommand{\ourdatasetinit}[0]{\textsc{MustardSauce}}

\newcommand{\ourdatasetInvalid}[0]{\ourdataset-\texttt{invalid}}
\newcommand{\ourdatasetRandom}[0]{\ourdataset-\texttt{random}}
\newcommand{\ourdatasetValid}[0]{\ourdataset-\texttt{valid}}
\newcommand{\ourdatasetValidTest}[0]{\ourdataset-\texttt{test}}
\newcommand{\ourdatasetTotal}[0]{\ourdataset-\texttt{tt}}

\newcommand{\ourdtInvalid}[0]{\texttt{in}}
\newcommand{\ourdtRandom}[0]{\texttt{ra}}
\newcommand{\ourdtValid}[0]{\texttt{va}}
\newcommand{\ourdtTotal}[0]{\texttt{tt}}
\newcommand{\mathlibTrain}[0]{\texttt{mt}}
\newcommand{\gsmekTrain}[0]{\texttt{gt}}

\makeatletter
\def\@fnsymbol#1{\ensuremath{\ifcase#1\or \dagger\or \ddagger\or
   \mathsection\or \mathparagraph\or \|\or **\or \dagger\dagger
   \or \ddagger\ddagger \else\@ctrerr\fi}}
    \makeatother

\title{
Mustard: Mastering {Uniform} Synthesis of Theorem and Proof Data
}

\author{\\\textbf{Yinya Huang}\textsuperscript{1,6} ~~
\textbf{Xiaohan Lin}\textsuperscript{2} ~~
\textbf{Zhengying Liu}\textsuperscript{3}\thanks{Corresponding author.} ~~
\textbf{Qingxing Cao}\textsuperscript{2}$^{\dag}$ ~~
\textbf{Huajian Xin}\textsuperscript{2} \\
\textbf{Haiming Wang}\textsuperscript{2} ~~
\textbf{Zhenguo Li}\textsuperscript{3} ~~
\textbf{Linqi Song}\textsuperscript{1,6}$^{\dag}$ ~~
\textbf{Xiaodan Liang}\textsuperscript{2,4,5}$^{\dag}$ \\
$^1$City University of Hong Kong ~~
$^2$Shenzhen Campus of Sun Yat-sen University \\
$^3$Huawei Noah's Ark Lab ~~
$^4$DarkMatter AI Research ~~
$^5$MBZUAI ~~
$^6$CityUSRI \\
\tt yinya.huang@hotmail.com,
\tt linxh55@mail2.sysu.edu.cn, \\
\tt \{liuzhengying2, Li.Zhenguo\}@huawei.com,
\tt caoqx8@sysu.edu.cn, \\
\tt linqi.song@cityu.edu.hk,
\tt xdliang328@gmail.com
}

\iclrfinalcopy 
\begin{document}

\maketitle

\begin{abstract}
%
Recent large language models (LLMs) have witnessed significant advancement in various tasks, including mathematical reasoning and theorem proving. As these two tasks require strict and formal multi-step inference, they are appealing domains for exploring the reasoning ability of LLMs but still face important challenges. Previous studies such as Chain-of-Thought (CoT) have revealed the effectiveness of intermediate steps guidance. However, such step-wise annotation requires heavy labor, leading to insufficient training steps for current benchmarks. 
To fill this gap,
this work introduces \ourmethod, a data generation framework that \textbf{\underline{m}}asters \textbf{\underline{u}}niform \textbf{\underline{s}}ynthesis of \textbf{\underline{t}}heorem \textbf{\underline{a}}nd p\textbf{\underline{r}}oof \textbf{\underline{d}}ata of high quality and diversity.
\ourmethod\ synthesizes data in three stages: (1) It samples a few mathematical concept seeds as the problem category. (2) Then, it prompts a generative language model with the sampled concepts to obtain both the problems and their step-wise formal solutions. (3) Lastly, the framework utilizes a proof assistant (e.g., Lean Prover)
to filter the valid proofs.
With the proposed \ourmethod, we present a theorem-and-proof benchmark \ourdataset\ with 5,866 valid data points.
Each data point contains an informal statement, an informal proof, and a translated formal proof that passes the prover validation.
We perform extensive analysis and 
demonstrate that \ourmethod\ generates validated high-quality step-by-step data. 
We further apply the \ourdataset\ for {fine-tuning smaller language models}. 
\update{
The fine-tuned Llama 2-7B achieves a 15.41\% average relative performance gain in automated theorem proving, 
and 8.18\% in math word problems.\footnote{Codes and data are available at: \href{https://github.com/Eleanor-H/MUSTARD}{https://github.com/Eleanor-H/MUSTARD}.}
}
\end{abstract}

\vspace{-3mm}
\section{Introduction}
\label{sec:intro}
\vspace{-1mm}
Large language models (LLMs) \citep{DBLP:journals/corr/abs-2303-08774,gpt35} have shown promising reasoning capabilities in various domains,
including math word problem
and theorem proving \citep{DBLP:journals/corr/abs-2110-14168,DBLP:conf/nips/HendrycksBKABTS21,DBLP:conf/iclr/ZhengHP22,DBLP:conf/iclr/WuJBG21}. These two tasks, which require strictly and successively multi-step inference, have become appeal domains to evaluate and develop LLMs' ability in complex reasoning. Recent works progress LLMs in solving math problems mainly through two techniques. The first is the chain-of-thoughts (CoT) prompting~\citep{DBLP:conf/nips/Wei0SBIXCLZ22,DBLP:conf/nips/KojimaGRMI22,DBLP:conf/iclr/0002WSLCNCZ23}, which provides step-by-step solutions to the LLMs. The second is to leverage the LLMs' ability in code generation to generate formalized languages and utilize external solvers to obtain strict inference results~\citep{DBLP:conf/nips/WuJLRSJS22,DBLP:conf/iclr/JiangWZL0LJLW23,DBLP:journals/corr/abs-2009-03393,DBLP:conf/iclr/HanRWAP22,DBLP:conf/iclr/PoluHZBBS23}. Both techniques rely on step-wise annotation to improve LLMs' performance and interpretability on the math problem.

\begin{figure}[!t]
    \centering
    \vspace{-12mm}
    \includegraphics[width=\textwidth]{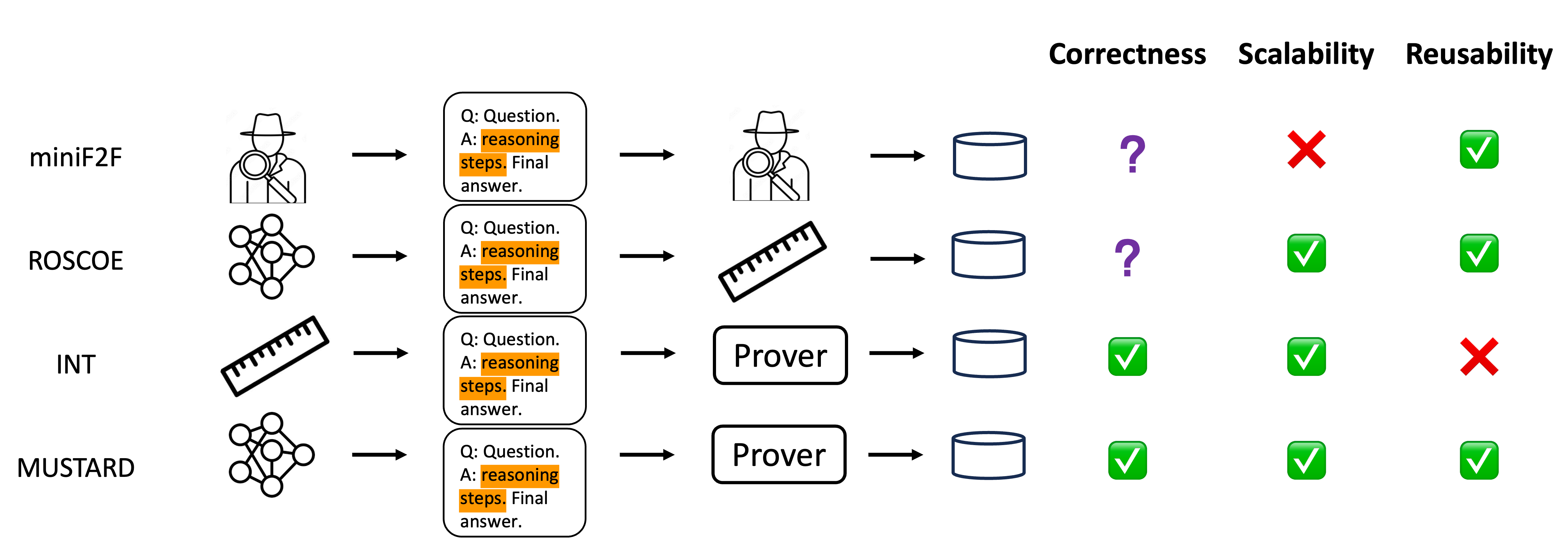}
    \vspace{-5mm}
    \caption{
    A comparison of methods of synthesizing and validating intermediate reasoning steps.
    }
    \vspace{-1mm}
    \label{fig:intuition}
\end{figure}

Correct intermediate steps
are crucial for LLMs to perform complex reasoning. 
However, high-quality step-wise annotations are hard to obtain, and 
Figure~\ref{fig:intuition} demonstrates a few representative works.
Previous works such as 
miniF2F \citep{DBLP:conf/iclr/ZhengHP22} resorts to manual annotation and validation to obtain high-quality 
step-wise labels.
However, manual annotation 
requires heavy labor of knowledgeable experts, resulting in an extremely small-scale dataset. Manual checking also does not guarantee the correctness of data as the labelers would make mistakes in labeling. 
On the other hand, generating data with rule-based checking such as 
ROSCOE \citep{DBLP:conf/iclr/GolovnevaCPCZFC23} can produce large-scale reasoning data.
Given that the generated data are more friendly and readable for human beings, the correctness of the reasoning is not guaranteed by those rules. 
Moreover, 
another line of work such as 
INT \citep{DBLP:conf/iclr/WuJBG21} performs rule-based synthesis to generate validated proofs, which are both correct and large-scale. However, the data are brutally synthesized so that many generated proofs lack actual meaning. 
Therefore, we need a more efficient way to generate mathematical data that are large-scale, with accurate intermediate steps, and also meaningful mathematical knowledge to human beings. 

To fill this gap, 
we propose \ourmethod, a data generation framework that 
{uniformly} synthesizes large-scale and high-quality mathematical data 
by combining the advantages of LLMs in verbalization and formal theorem provers in rigorous data validation.
Specifically, \ourmethod\ first samples a few mathematical concepts from a predefined list and prompts an LLM to generate a related question described in natural language. 
Then, it applies the LLM to generate the corresponding solution in both natural and formal language.
Given the generated solution, \ourmethod\ further validates them using a theorem prover. The passed one is considered to be correct and is a high-quality data point. The invalid one on the other hand is considered to be a challenging sample, which will be further combined with the error messages to prompt the LLM for a solution revision, and added as a challenging data point. 

By applying the proposed \ourmethod\, one can obtain large amounts of problems and theorems with desired mathematical concepts and domains. Eventually, we build a mathematical dataset with validated informal and formal solutions, named \ourdataset\ (\ourmethod\ resource). 


We conduct extensive data analysis and experiments on the generated \ourdataset. 
Through deep inspection of the data, we find that \ourmethod\ generates interesting and reasonable math problems by creatively combining two mathematical concepts, and \ourdataset\ is diverse and has a high proportion of difficult data.
We also observe that the prover is consistent with human evaluation, where humans usually consider a validated solution to have a higher quality than those without a formal validation process.
Lastly, we fine-tune smaller-scale language models
on \ourdataset.
The fine-tuned Llama 2-7B achieves improvements by 20.9\% on zero-shot inference on GSM8K and achieves 8.7 of pass@1 on mathlib.
These results demonstrate the effectiveness of \ourdataset\ in improving the mathematical reasoning capabilities of language models.

\vspace{-1mm}

The contributions of this paper are summarized as follows:
\begin{enumerate}
    \vspace{-1mm}
    \item We propose a novel framework \ourmethod\ that can generate high-quality mathematical data (both informal and formal) with an interplay between generative language model and theorem prover assistants.
    \vspace{-1mm}
    \item We release the \ourdataset,
    which contains both math word problems and theorem-proving problems spanning over four educational levels. 
    Each sample has corresponding informal and formal solutions. 
    \vspace{-1mm}
    \item We conduct extensive analysis and experiments on the generated data, demonstrating their quality, diversity, and effectiveness in improving language models' mathematical reasoning performance.
\end{enumerate}




\vspace{-2mm}
\section{Related Works}  
\label{sec:related}
\vspace{-3mm}
\paragraph{Large Language Models for Mathematical Reasoning}
The growing generative language models \citep{DBLP:conf/nips/BrownMRSKDNSSAA20,gpt35,DBLP:journals/corr/abs-2303-08774} show compelling potential for solving mathematical problems
both in natural language proofs \citep{DBLP:journals/corr/abs-2303-08774,zheng2023progressive,xiong2024dqlore}
and in formal languages with theorem provers \citep{DBLP:journals/corr/abs-2009-03393,DBLP:conf/iclr/HanRWAP22,zheng2023lyra,DBLP:conf/iclr/PoluHZBBS23}. 
On the other hand, some works explore using language models to automatically translate natural language proofs into formal ones given few-shot demonstrations \citep{DBLP:conf/nips/WuJLRSJS22,DBLP:conf/iclr/JiangWZL0LJLW23,DBLP:journals/corr/abs-2309-04295,wang2024legoprover}. 
Chain-of-though reasoning \citep{DBLP:conf/nips/Wei0SBIXCLZ22,DBLP:conf/nips/KojimaGRMI22,DBLP:conf/iclr/0002WSLCNCZ23} is demonstrated beneficial for the LLMs to derive correct answers.
However, some recent works \citep{DBLP:conf/iclr/Saparov023,DBLP:conf/iclr/GolovnevaCPCZFC23} observe that the intermediate reasoning steps can be inconsistent.
%
%
This paper proposes a data generation framework that taps the comprehensive mathematical reasoning capabilities of large language models.
It generates mathematical reasoning problems with informal and formal solutions that are step-wise validated by a formal theorem prover.
With the framework, we obtain high-quality mathematical data.

\vspace{-3mm}
\paragraph{Synthesizing Mathematical Data}
Obtaining large-scale high-quality mathematical data is a long-standing challenge. 
Previous data relies on well-trained annotators to hand-craft and review the formal proofs \citep{DBLP:conf/iclr/ZhengHP22}, which is time and labour-consuming and results in a small data scale. 
\cite{DBLP:conf/nips/WangD20} and \cite{xiong-etal-2023-trigo} construct a neural generator for data synthesis, but it still requires the intervention of human-written data.
Besides, \cite{DBLP:conf/iclr/WuJBG21} explore using a theorem generator to automatically generate formal proofs with rules.
However, the rule-based generation depends on given axioms in specified orders. As a result, the generated data is restricted to a few domains.
%
On the other hand, recent works demonstrate the effectiveness of distilling knowledge from large language models \citep{DBLP:conf/naacl/WestBHHJBLWC22,DBLP:conf/acl/YuanCFGSJXY23,DBLP:journals/corr/abs-2308-06259},
and some of them \citep{DBLP:conf/acl/WangKMLSKH23,DBLP:journals/corr/abs-2304-12244} explore data evolution by properly prompting the language models.
%
The proposed framework explores eliciting mathematical knowledge from large language models to achieve diverse and large-scale mathematical data. 
In this framework, an interplay between the language model and a formal proof assistant controls the quality and difficulties of data.
Using the proposed framework, we collect a large-scale mathematical dataset that contains diverse and multiple-difficulty math questions with high-quality solutions. 

\vspace{-3mm}
\section{\ourmethod}  
\begin{figure}[!t]
    \centering
    \vspace{-3mm}
    \includegraphics[width=\textwidth]{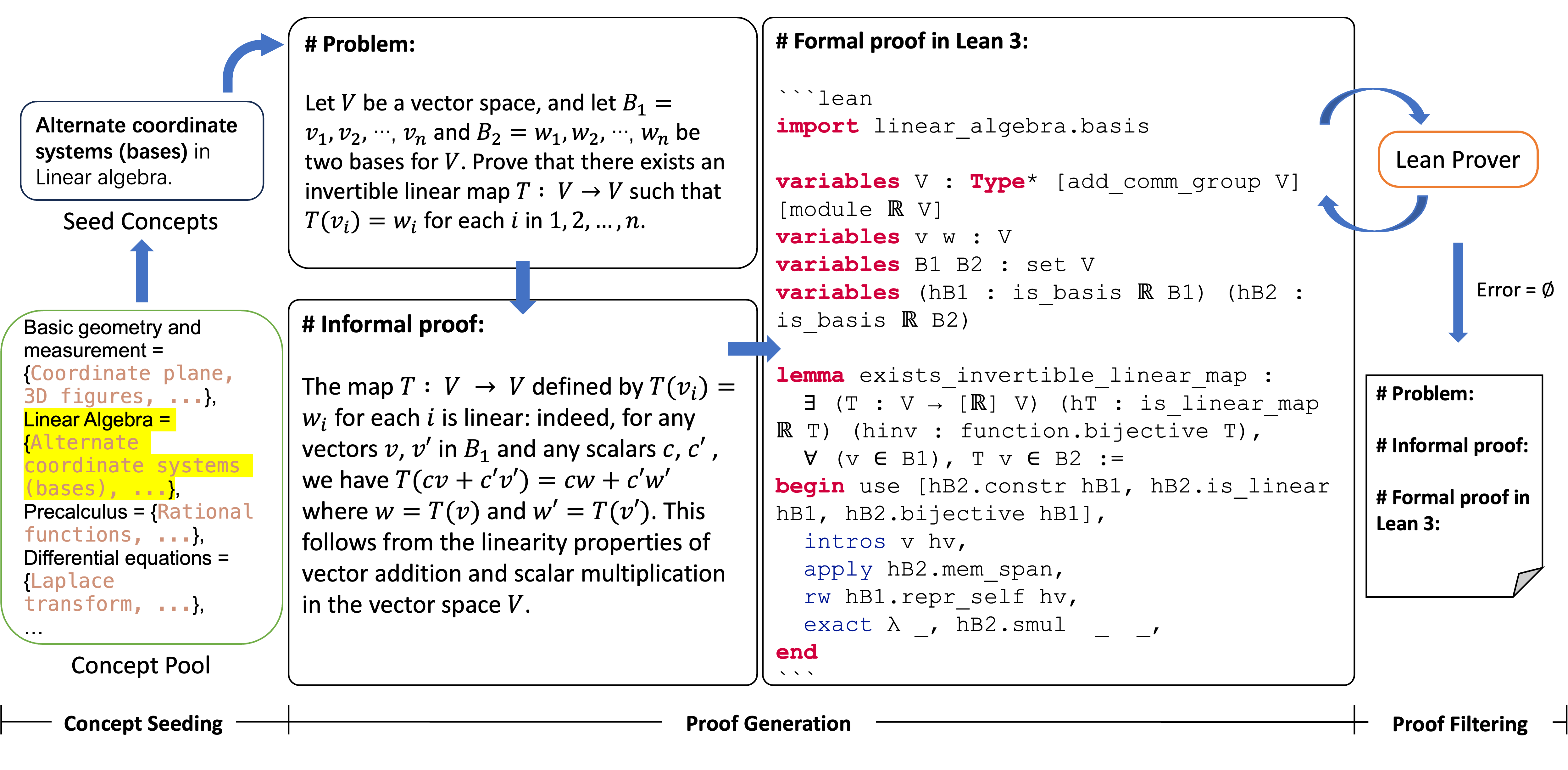}
    \vspace{-2mm}
    \caption{
    Overview of \ourmethod.
    The mathematical theorem (indicated by ``\# Problem:'') is created according to the randomly selected seed concepts.
    Following the theorem, a corresponding informal proof is generated, 
    which is then translated into a formal proof in Lean 3. 
    The formal proof is passed into Lean Prover for quality feedback, 
    and according to the error messages the formal proof is revised and improved. 
    The proofs that pass the Lean Prover are regarded as high-quality data points and collected.
    }
    \vspace{-2mm}
    \label{fig:framework}
\end{figure}

\vspace{-3mm}
In this work, we aim to obtain large-scale mathematical data with multi-step annotations and propose \ourmethod\ to generate diverse and high-quality math and theorem-proving problems with multi-step informal and formal solutions. 
As shown in Figure~\ref{fig:framework}, \ourmethod\ consists of three stages. 
In the first concept seeding stage, \ourmethod\ samples a set of math concepts as the problem domain. 
Then in the second solution generation stage, it generates the concept-related problem and solution by prompting an LLM. 
In the third stage, a theorem prover is used to validate the generated solution. If the solution can not pass the prover, the error message is returned to the second stage for another turn of solution generation.
Through interaction between the LLM and a formal proof assistant, \ourmethod\ can generate diverse and high-quality data
that contains both informal and formal solutions. 
We describe the details of each stage in this section.

\vspace{-3mm}
\subsection{Concept Seeding}
\vspace{-3mm}
We first define and build a mathematical concept pool 
that covers as complete sub-subjects in mathematics and educational levels as possible. 
Specifically, we collect all math courses on the Khan Academy website\footnote{\href{https://www.khanacademy.org/math}{https://www.khanacademy.org/math}}, 
the large-scale online educational platform.
%
The resulting pool includes concepts in four educational levels: elementary school, middle school, high school, and higher education.
Each educational level has 5 to 9 math domains, covering different types of math problems such as algebra and geometry. Each domain contains subdivided mathematical concepts to inspect different mathematical abilities like polynomial arithmetic or factorization. 
Concept statistics and detailed concepts in each domain are demonstrated in Appendix~\ref{app:concepts}.

Given the concept pool, for each educational level, \ourmethod\ 
uniformly samples 1 or 2 concepts from all domains as seeds,
and then generates mathematical problems that cover the concepts.
In particular, given an educational level, taking 2 concepts from different subjects challenges the model to generate problems that join diverse domains while keeping the problems reasonable.

\subsection{Proof Generation} 
\vspace{-2mm}
Given the sampled mathematical concepts, \ourmethod\ generates math problems and their corresponding solutions.
Specifically, \ourmethod\ leverages the capability of LLMs in generating natural language and code and prompts an LLM to generate the problem statement, its natural language solution, and a formal solution written in Lean. 
As a result, 
the LLM needs to complete the following three tasks:
(T1) Generating a math problem that relates to the given concepts; 
(T2) Solving the math problem with a natural language proof; 
(T3) Performing auto-formalization to translate the written natural language proof into a formalized proof. \update{In this work, we use GPT-4~\cite{DBLP:journals/corr/abs-2303-08774} as the LLM for proof generation.}

We intend to generate a problem based on educational level, math domains, and concepts. 
Considering that mathematical problems include proof and calculation, we also introduce the question types into the prompt for generating theorem proving and word problems respectively. 
Moreover,
we do not include any exemplars or other manual interventions except for the sampled concepts. We intend to avoid potential biases brought by the concepts inside the exemplars and achieve a more diverse generation.
%
%
The prompt template is shown as the following:
\begin{table}[!h]
    \scriptsize
    \vspace{-2mm}
    \begin{tabular}{p{\textwidth}}
         \texttt{You are a math expert. Now please come up with a math problem according to the following requirements. The math problem should contain a question part (indicated by ``Problem: ''), a corresponding solution in natural language (indicated by ``Informal proof:''), and a translated formal solution in Lean 3 (indicated by ``Formal proof in Lean 3:''). Please note that the informal proof and the formal proof need to be identical.}\\ \\
        \texttt{Please create a [QUESTION TYPE] in the level of [EDUCATIONAL LEVEL] based on the following knowledge point(s): \*[CONCEPT]\* in [DOMAIN]; \*[CONCEPT]\* in [DOMAIN].} \\ \\
        \texttt{You must respond in the following format: } \\ \\
        \texttt{\# Problem: ...} \\ \\ 
        \texttt{\# Informal proof: ...} \\ \\
        \texttt{\# Formal proof in Lean 3: ...} \\ 
    \end{tabular}
    \vspace{-2mm}
\end{table}

The ``\texttt{[]}'' indicates the placeholders for the corresponding
question type, educational level, concepts, and domains. 
Multiple concepts are separated by ``;''. 
We retrieved the text after the ``Problem:'', ``Informal proof:'' and ``Formal proof in Lean 3:'' as the generated sample.

\vspace{-3mm}
\subsection{Proof Filtering}
\vspace{-2mm}
In the proof-filtering stage, 
\ourmethod\ interacts with the Lean Prover \citep{DBLP:conf/cade/MouraKADR15} to obtain validation messages of the proof steps, which guides data revision and filtering. 
Specifically, after a formal solution is passed to the Lean Prover and
if the prover returns no error message, 
the corresponding data point is collected into the valid dataset. 
Otherwise, \ourmethod\ collects the error messages from the prover and prompts the language model to revise the invalid solution. 
To help the language model locate the incorrect lines described in the error messages, we also add a line number at the beginning of each line in the formal solutions. 
The verification and self-refinement
are performed in multiple rounds until LLM generates a valid solution. We use the number of rounds to measure the difficulty of the generated sample, assuming a difficult problem is hard to solve by an LLM and requires more rounds of correction.
The prompt template of a single round of correction is demonstrated as follows, and the complete prompt template is shown in Table~\ref{tab:prompt_proof_filtering} in Appendix~\ref{app:proof_filtering}:
\begin{table}[!h]
    \scriptsize
    \vspace{-2mm}
    \begin{tabular}{p{\textwidth}}
        \texttt{\# Formal proof (c) in Lean 3: } \\ 
        \texttt{```lean} \\ 
        \texttt{line 1 <code> } \\ 
        \texttt{line 2 <code> } \\ 
        \texttt{line 3 <code> } \\ 
        \texttt{... } \\ 
        \texttt{```} \\ \\
        \texttt{\# Error messages for Formal proof (c) from Lean Prover: } \\ 
        \texttt{<error messages>}\\ \\
    \end{tabular}
    \vspace{-4mm}
\end{table}

\vspace{-3mm}
\section{Experiments}
\label{sec:exp}
\vspace{-2mm}

\vspace{-2mm}
\subsection{Case Study}
\vspace{-2mm}
We first inspect the data points generated by \ourmethod.
Table~\ref{tab:main_good_case_3} shows a generated math problem in which \ourmethod\ creatively combines two mathematical concepts and constructs a reasonable question.
The generated question includes knowledge from both concepts. 
It is suggested that \ourmethod\ can join the concepts and construct a reasonable question. 
%
Furthermore, Table~\ref{tab:main_good_case_5} demonstrates a case that \ourmethod\ provides solid and comprehensive solutions in both natural language and Lean. 
Although the formal proof is long, it is consistent and passes the prover's validation. 
It is demonstrated that \ourmethod\ can generate long valid solutions.


\vspace{-1mm}
\begin{table}[!t]
    \vspace{-4mm}
    \caption{An informal statement generated by \ourmethod.
    The complete data point is demonstrated in Table~\ref{tab:good_case_3} in Appendix~\ref{app:case_study}.
    }
    \label{tab:main_good_case_3}
    \small
    \centering
    \vspace{-2mm}
    \begin{tabular}{
        p{\textwidth}
    }
        \toprule
        \textbf{Question Type: Theorem Proving.\quad Educational Level: Middle School.\quad k=2. } \\
        \midrule
        \textbf{Concept(s): } \textit{Geometry} in 8th grade; \textit{Algebraic expressions} in Algebra basics. \\
        \midrule
        \textbf{Informal Statement} \\
        Given a rectangle ABCD where AB is x + 5 and AD is 2x - 3. Prove that the area of the rectangle ABCD is $(2x^2 + 7x - 15)$ square units. \\ 
        \bottomrule
    \end{tabular}
    \vspace{-2mm}
\end{table}

\begin{table}[!t]
    \vspace{-1mm}
    \caption{A data point generated by \ourmethod.}
    \label{tab:main_good_case_5}
    \small 
    \centering
    \vspace{-2mm}
    \begin{tabular}{
        p{\textwidth}
    }
        \toprule
        \textbf{Question Type: Word Problem.\quad Educational Level: Higher Education.\quad k=1. } \\
        \midrule
        \textbf{Concept(s): } \textit{Series} in Integral Calculus. \\
        \midrule
        \textbf{Informal Statement} \\
        Prove that the integral of the series $\sum$ (from $n=1$ to infinity) of $\frac{-1^{(n+1)}}{n}$ equals $\ln(2)$. \\ \\
        \textbf{Informal Proof} \\
        We start by recognizing the series $\sum$ (from $n=1$ to infinity) of $\frac{-1^{(n+1)}}{n}$ as the alternating harmonic series, which is known to converge to $\ln(2)$.  \\ \\
        \textbf{Formal Statement and Proof} \\
        \texttt{\textbf{\textcolor{purple}{import}} data.real.basic} \\
		\texttt{\textbf{\textcolor{purple}{import}} analysis.special\_functions.pow} \\ \\

		\texttt{\textbf{\textcolor{purple}{open}} real} \\ \\

		\texttt{\textbf{\textcolor{purple}{lemma}} integral\_of\_series : } \\
		\texttt{  $\int$ x \textbf{\textcolor{purple}{in}} 0..1, (ln (1 + x)) = $\sum^\prime$ (n : $\mathbb{N}$), ((-1)\textasciicircum(n+1) / (n + 1)) :=} \\
		\texttt{\textbf{\textcolor{purple}{begin}}} \\
		\texttt{  \textbf{\textcolor{purple}{have}} h\_der : $\forall$ x $\in$ set.Icc (0 : $\mathbb{R}$) 1, deriv ($\lambda$ x, (ln (1 + x))) x = 1 / (1 + x),} \\
		\texttt{  \{ \tacticBlue{intros} x hx,} \\
		\texttt{    \textbf{\textcolor{purple}{have}} h\_pos : 0 < 1 + x := \textbf{\textcolor{purple}{by}} \tacticBlue{nlinarith},} \\
		\texttt{    \tacticBlue{exact} deriv\_ln\_one\_add h\_pos \},} \\ \\
		  
		\texttt{  \textbf{\textcolor{purple}{have}} h\_int : interval\_integral ($\lambda$ x, 1 / (1 + x)) 0 1 volume = $\sum^\prime$ (n : $\mathbb{N}$), ((-1)\textasciicircum(n+1) / (n + 1)),} \\
		\texttt{  \{ \textbf{\textcolor{purple}{have}} h\_frac : $\forall$ (n : $\mathbb{N}$), $\int$ x \textbf{\textcolor{purple}{in}} 0..1, x\textasciicircum n = 1 / (n + 1),} \\
		\texttt{    \{ \tacticBlue{intro} n,} \\
		\texttt{      \textbf{\textcolor{purple}{calc}} $\int$ x \textbf{\textcolor{purple}{in}} 0..1, x\textasciicircum n = [x\textasciicircum(n+1) / (n+1)] | 0..1 : integral\_pow n} \\
		\texttt{      ... = 1 / (n + 1) : \textbf{\textcolor{purple}{by}} \{ \tacticBlue{rw} integral\_interval, \tacticBlue{simp}  \} \},} \\
		\texttt{    \tacticBlue{rw} [interval\_integral.integral\_of\_le, h\_frac],} \\
		\texttt{    \tacticBlue{simp},} \\
		\texttt{    \tacticBlue{linarith} \},} \\ \\
		  
		\texttt{  \textbf{\textcolor{purple}{have}} h\_eq : $\int$ x \textbf{\textcolor{purple}{in}} 0..1, (ln (1 + x)) = $\int$ x \textbf{\textcolor{purple}{in}} 0..1, 1 / (1 + x), } \\
		\texttt{  \{ \tacticBlue{congr}, \tacticBlue{ext}, \tacticBlue{exact} h\_der x (set.mem\_Icc.mpr ⟨\textbf{\textcolor{purple}{by}} \tacticBlue{nlinarith}, \textbf{\textcolor{purple}{by}} \tacticBlue{nlinarith}⟩) \},} \\ \\
		  
		\texttt{  \tacticBlue{rw} [h\_eq, h\_int],} \\
		\texttt{\textbf{\textcolor{purple}{end}}} \\ 
        \bottomrule
    \end{tabular}
    \vspace{-3mm}
\end{table}

\vspace{-2mm}
\subsection{Human Evaluation}  
\label{sec:human_eval}
\vspace{-2mm}
To further explore the quality of the data generated by \ourmethod, we recruit professionals who have expertise in mathematics and the Lean language to perform a sanity check on the data points. 
We randomly select \update{200} data points from the generated data, \update{100} of which pass the Lean Prover (Group Valid) and \update{100} of which do not (Group Invalid). 
The sanity check covers the four sections in each data point (i.e., informal statement, informal proof, formal statement, and formal proof), and includes factuality check and consistency check. 
Specifically, a high-quality data point should have a factually correct informal statement (D1) and a correct solution (D4). 
The formal statement and proof should be aligned with the informal descriptions (D5, D6). 
Moreover, the desired data point should meet the specified seed concepts (D2) and question type (D3).
The six inspection dimensions and their requirements are demonstrated in Table~\ref{tab:human_dims}.
A data point is scored 1 in a dimension if it meets the requirement, otherwise, it gets 0. 
\vspace{-1mm}
\begin{table}[!h]
    \scriptsize
    \centering
    \vspace{-4mm}
    \caption{
        Inspection dimensions and requirements in human evaluation.
        IS: Informal Statement. 
        IP: Informal Proof.
        FS: Formal Statement.
        FP: Formal Proof.
        RT: Reasoning Type. 
        \update{Significant $p<0.005$ are marked with \textbf{bold}.}
    }
    \vspace{-2mm}
    \label{tab:human_dims}
    \begin{tabular}{
        lp{.45\textwidth}|ccc
    }
    \toprule
        Inspection Dimension & Requirement & Valid & Invalid & $p$-value \\
    \midrule
    (D1) IS Correctness & 
            \makecell[{{{>{\raggedright\arraybackslash}p{.45\textwidth}}}}]{\textit{Whether the informal statement is factually correct. }}
            & \update{93.50} & \update{83.50} & \update{\textbf{0.00167}} \\
    (D2) IS Relevance & 
            \makecell[{{{>{\raggedright\arraybackslash}p{.45\textwidth}}}}]{\textit{Whether the informal statement is relevant to each seed concept.}} 
            & \update{87.50} & \update{92.50} & \update{0.09604} \\
    (D3) RT Classification & 
            \makecell[{{{>{\raggedright\arraybackslash}p{.45\textwidth}}}}]{\textit{Whether the informal statement is of the required question type.}} 
            & \update{67.00} & \update{68.50} & \update{0.74903} \\
    (D4) IP Correctness & 
            \makecell[{{{>{\raggedright\arraybackslash}p{.45\textwidth}}}}]{\textit{Whether the informal proof correctly solves the informal statement.}} 
            & \update{88.50} & \update{73.50} & \update{\textbf{0.00012}} \\
    (D5) IS-FS Alignment & 
            \makecell[{{{>{\raggedright\arraybackslash}p{.45\textwidth}}}}]{\textit{Whether the informal statement and the formal statement describe the same problem and are aligned with each other.}} 
            & \update{74.00} & \update{66.50} & \update{0.10138} \\
    (D6) IP-FP Alignment & 
            \makecell[{{{>{\raggedright\arraybackslash}p{.45\textwidth}}}}]{\textit{Whether the informal proof and the formal proof describe the same solution and have aligned proof steps.}} 
            & \update{72.00} & \update{54.00} & \update{\textbf{0.00018}} \\
    \bottomrule
    \end{tabular}
    \vspace{-2mm}
    \label{tab:dim}
\end{table}

\begin{table}[!t]
    \caption{
        Maj1@1 results on GSM8K \update{(G) and MATH (M)}. Zero: Zero-shot. Few: Few-shot. 
        \textgreater\ denotes a fine-tuning step. 
        \ourdtInvalid: \ourdatasetInvalid.
        \ourdtRandom: \ourdatasetRandom.
        \ourdtValid: \ourdatasetValid.
        \update{\ourdtTotal: \ourdatasetTotal.}
        \gsmekTrain: GSM8K training split.
    }
    \vspace{-2mm}
    \label{MWP}
    \scriptsize
    \centering
    \begin{tabular}{
            p{1.5cm}p{.8cm}p{.8cm}p{.8cm}p{.8cm}
            p{1.5cm}p{.8cm}p{.8cm}p{.8cm}p{.8cm}
        }
        \toprule
        MODEL                   & Zero (G) & Few (G) & \update{Zero (M)} & \update{Few (M)}  
        & MODEL                 & Zero (G) & Few (G) & \update{Zero (M)} & \update{Few (M)} \\ 
        \midrule
        \multicolumn{6}{l}{\textit{Baselines}}  \\
        GPT2-large                              & 3.4 & 5.1 & \update{0.6} & \update{1.0} & GPT2-large \textgreater\ \gsmekTrain  & 14.6 & 17.4 & \update{4.6} & \update{6.8} \\
        Llama 2-7B                              & 7.2 & 12.8 & \update{2.0} & \update{2.6} & Llama 2-7B \textgreater\ \gsmekTrain & 24.5 & 28.2 & \update{10.4} & \update{12.6} \\
        \midrule
        \multicolumn{6}{l}{\textit{Fine-tuning}}   \\
        \update{GPT2-large\ \textgreater\ \ourdtTotal}  & 
                {4.2} & 6.8 & 1.4 & 2.4 & 
            \update{ GPT2-large \textgreater\ \ourdtTotal\ \textgreater\ \gsmekTrain}  & 
                {16.1} & 18.9 & 4.8 & 8.0 \\
        GPT2-large\ \textgreater\ \ourdtInvalid  & 
                {3.9} & 6.4 & 1.2 & 2.0 &
            GPT2-large \textgreater\ \ourdtInvalid\ \textgreater\ \gsmekTrain  & 
                {15.4} & 17.7 & 4.4 & 7.6 \\
        GPT2-large\ \textgreater\ \ourdtRandom   & 
                {4.1} & 6.7 & 1.4 & 2.2 &
            GPT2-large \textgreater\ \ourdtRandom\ \textgreater\ \gsmekTrain   & 
                {15.7} & 18.5 & 4.8 & 7.8 \\
        GPT2-large\ \textgreater\ \ourdtValid    & 
                {4.6 (+12.20\%)} & 7.0 (+4.48\%) & 1.8 (+28.57\%) & 2.8 (+27.27\%) & 
            GPT2-large \textgreater\ \ourdtValid\ \textgreater\ \gsmekTrain    & 
                {16.5 (+5.10\%)} & 20.1 (+8.65\%) & 5.6 (+16.67\%) & 8.4 (+7.69\%) \\
        \update{Llama 2-7B\ \textgreater\ \ourdtTotal}  & 
                {9.6} & 16.0 & 3.2 & 3.8 & 
            \update{Llama 2-7B \textgreater\ \ourdtTotal\ \textgreater\ \gsmekTrain}  & 
                {27.4} & 31.5 & 13.0 & 14.4 \\
        Llama 2-7B\ \textgreater\ \ourdtInvalid  & 
                {9.1} & 14.9 & 2.4 & 3.2 & 
            Llama 2-7B \textgreater\ \ourdtInvalid\ \textgreater\ \gsmekTrain & 
                {26.9} & 30.3 & 12.4 & 13.8 \\
        Llama 2-7B\ \textgreater\ \ourdtRandom   & 
                {9.5} & 15.4 & 3.0 & 3.6 & 
            Llama 2-7B \textgreater\ \ourdtRandom\ \textgreater\ \gsmekTrain  & 
                {27.1} & 30.7 & 12.6 & 14.2 \\
        Llama 2-7B\ \textgreater\ \ourdtValid   & 
                {10.3 (+8.42\%)} & 16.9 (+9.74\%) & 3.2 (+6.67\%) & 4.2 (+16.67\%) & 
            Llama 2-7B \textgreater\ \ourdtValid\ \textgreater\ \gsmekTrain    & 
                {27.9 (+2.95\%)} & 32.5 (+5.86\%) & 13.8 (+9.52\%) & 15.0 (+5.63\%) \\
        \bottomrule
    \end{tabular}
\end{table}

The accuracies of Group Valid and Group Invalid in each dimension are demonstrated in Table~\ref{tab:human_dims}.
We also report the corresponding $p$-value in each dimension. 
(D4) and (D6) show significant differences in accuracy between the two groups.
The results indicate that high-quality data points have significantly better auto-formalization results.
As a result, given the validation of formal proofs by the Lean Prover, the data points have guaranteed high-quality informal proofs. 
\update{Moreover, (D1) also shows significance with the inspected data scaled up.}
The differences in statement alignment (D5) and informal statement relevance (D2) of the two groups are less significant. 
%
Furthermore, no significant differences are observed in question type classification (D3),
\update{which indicates that Lean Prover validation in \ourmethod\ does not significantly influence the classification.}
%
%
Overall, the human evaluation results suggest that formally validated data have significantly higher quality. 


\vspace{-2.7mm}
\subsection{Data Quality by Downstream Application} 
\vspace{-2mm}
To evaluate the impact of \ourdataset\ on enhancing mathematical reasoning abilities, we use the data to fine-tune smaller-scale language models and evaluate them on math word problems (MWP) and automated theorem proving (ATP).
Specifically,
given all the generated data, 
we first have 5,866 valid data points that pass the Lean Prover.
We denote this subset \ourdatasetValid.
We then extract the same number of invalid data points as the subset of \ourdatasetInvalid,
and extract an equal size of random subset \ourdatasetRandom. 
Each \ourdataset\ subset contains 5,866 data points. 
Moreover, we randomly split \ourdatasetValid\ into 4,866 training data, 500 validation data, and 500 test data for benchmarking model performances on the dataset. 
We denote the test set as \ourdatasetValidTest.
\update{Furthermore, we also test on the entire generated data set with 28,316 data points, which we denote \ourdatasetTotal.}


\vspace{-.5mm}
We employ LoRA~\citep{hu2021lora} for fine-tuning the open-source GPT2-large~\citep{radford2019language}, Llama 2-7B 
\update{and Llama 2-70B}~\citep{DBLP:journals/corr/abs-2307-09288}
on each \ourdataset\ subset.
Detailed model configuration and training procedure are described in Appendix~\ref{app:downstream}.
For the task of math word problems, we use
GSM8K \citep{DBLP:journals/corr/abs-2110-14168} \update{and MATH dataset \cite{DBLP:conf/nips/HendrycksBKABTS21}}\footnote{\update{Specifically, we use the released test split with 500 problems in PRM800K \citep{lightman2023let} which is demonstrated to be representative of the MATH test set as a whole.}} for evaluation.
For evaluating automated theorem proving, 
we use Mathlib\footnote{\href{https://github.com/leanprover-community/mathlib}{https://github.com/leanprover-community/mathlib}} and the miniF2F \citep{DBLP:conf/iclr/ZhengHP22} benchmark.
We also evaluate models on \ourdatasetValidTest\ after being fine-tuned on the \ourdatasetValid\ training split. 
Tables~\ref{MWP} and ~\ref{ATP} demonstrate the model performances. 
We also follow \cite{DBLP:conf/iclr/HanRWAP22} to ablate the fine-tuning steps
and demonstrate the results in Table~\ref{tab:ablation}.



In general, fine-tuning the models on \ourdataset\ improves the mathematical reasoning of the models. 
\update{
On average, we have an 18.15\% relative performance gain after fine-tuning with \ourdatasetValid\ compared with fine-tuning with \ourdatasetRandom\ in ATP (Table~\ref{ATP}) and 11.01\% in MWP (Table~\ref{MWP}). 
{The fine-tuned Llama 2-7B achieves average gains of 15.41\% and 8.18\% on ATP and MWP,
and the fine-tuned GPT 2-large 20.89\% and 15.41\%, respectively.}
Specifically, in ATP, the Llama 2-7B achieves significant performance gains of 16.00\% on both mathlib and miniF2F, while increasing 17.31\% on the \ourdatasetValidTest.
In MWP, the performance improvements are also consistent in two datasets and both zero-shot and few-shot inference. 
%
We further compare the results fine-tuned with \ourdatasetTotal\ and \ourdatasetValid.
We find that models fine-tuned with the entire generated data are inferior to models fine-tuned with MUSTARDSAUCE-valid. 
Although the increase in the amount of fine-tuned data makes the model perform better compared to fine-tuning on MUSTARDSAUCE-invalid and MUSTARDSAUCE-random, the model's performance still lags behind that of fine-tuning on smaller amounts but higher quality data. 
Therefore, our proposed framework that introduces the theorem prover is effective and beneficial.
Furthermore, complementary experimental results of larger Llama 2-70B are demonstrated in Table~\ref{tab:70b} in Appendix~\ref{app:res}. The results suggest that our method remains effective when fine-tuning a larger language model. 
}

\vspace{-3mm}
\begin{table}[!t]
    \vspace{-4mm}
    \caption{
        Pass@1 results on automated theorem proving tasks.
        \textgreater\ denotes a fine-tuning step. 
        \texttt{test}: \ourdatasetValidTest.
        \ourdtInvalid: \ourdatasetInvalid.
        \ourdtRandom: \ourdatasetRandom.
        \ourdtValid: \ourdatasetValid.
        \update{\ourdtTotal: \ourdatasetTotal.}
        \mathlibTrain: mathlib training split.
        Note that the reported results on \ourdatasetValidTest\ are obtained by only fine-tuning on the \ourdatasetValid\ training split. 
    }
    \vspace{-2mm}
    \label{ATP}
    \centering
    \scriptsize
    \begin{tabular}{
        lp{.8cm}p{.8cm}p{.8cm}
        lp{.8cm}p{.8cm}p{.8cm}
    }
        \toprule
        MODEL       & mathlib & miniF2F & \texttt{test} &  MODEL & mathlib & miniF2F & \texttt{test} \\
        \midrule
        \multicolumn{8}{l}{\textit{Baselines}}   \\
        GPT2-large & 0.0 & 0.0 & 0.0 & GPT2-large \textgreater\ \texttt{mt} & 5.6 & 2.9 & 8.6  \\
        Llama 2-7B & 0.0 & 0.0 & 0.0& Llama 2-7B \textgreater\ \texttt{mt} & 14.3 & 7.0 & 10.8 \\
        \midrule
        \multicolumn{8}{l}{\textit{Fine-tuning}} \\  
        GPT2-large \textgreater\ \ourdtInvalid & 2.0 & 0.0 & 6.0 & GPT2-large \textgreater\ \ourdtInvalid \textgreater\ \mathlibTrain  & 5.9 & 2.0 & 8.2  \\
        GPT2-large \textgreater\ \ourdtRandom & 3.0 & 1.2 & 7.0 & GPT2-large \textgreater\ \ourdtRandom \textgreater\ \mathlibTrain & 6.6 & 2.9 & 9.6  \\
        GPT2-large \textgreater\ \ourdtValid & 3.7 \update{(+23.33\%)} & 1.6 \update{(+33.33\%)} & 8.3 \update{(+18.57\%)} & GPT2-large \textgreater\ \ourdtValid \textgreater\ \mathlibTrain & 7.4 \update{(+12.12\%)} & 3.7 \update{(+27.59\%)} & 10.6 \update{(+10.42\%)} \\
        \update{Llama 2-7B \textgreater\ \ourdtTotal} & \update{8.3} & \update{2.6} & \update{11.7} 
            & \update{Llama 2-7B \textgreater\ \ourdtTotal \textgreater\ \mathlibTrain} & \update{15.1} & \update{7.0} & \update{13.6} \\
        Llama 2-7B \textgreater\ \ourdtInvalid & 5.8 & 1.2 & 8.6  & Llama 2-7B \textgreater\ \ourdtInvalid \textgreater\ \mathlibTrain & 11.6 & 5.7 & 12.6 \\
        Llama 2-7B \textgreater\ \ourdtRandom & 7.5 & 2.5 & 10.4 & Llama 2-7B \textgreater\ \ourdtRandom \textgreater\ \mathlibTrain & 14.7 & 6.6 & 13.2 \\
        Llama 2-7B \textgreater\ \ourdtValid & 8.7 \update{(+16.00\%)} & 2.9 \update{(+16.00\%)} & 12.2 \update{(+17.31\%)} & Llama 2-7B \textgreater\ \ourdtValid \textgreater\ \mathlibTrain & 15.7 \update{(+6.80\%)} & 7.8 \update{(+18.18\%)} & 14.4 \update{(+18.18\%)} \\
        \bottomrule
    \end{tabular}
    \vspace{-2mm}
\end{table}

\begin{table}[!t]
    \begin{minipage}{.45\linewidth}
        \vspace{-1mm}
        \caption{
            Ablation study of different fine-tuning settings.
            \texttt{test}: \ourdatasetValidTest.
            \ourdtValid: \ourdatasetValid.
            \mathlibTrain: mathlib training split.
        }
        \vspace{-2mm}
        \label{tab:ablation}
        \centering
        \small
        \begin{tabular}{
            lccc
        }
            \toprule
            MODEL &&& \texttt{test} \\
            \midrule
            GPT2-large\ \  \textgreater\ \ \ \texttt{va}                                     &&& 8.3 \\
            GPT2-large\ \  \textgreater\ \ \ \texttt{va}\ \ \ \textgreater\ \ \ \texttt{mt}  &&& 10.6 \\
            GPT2-large\ \  \textgreater\ \ \ \texttt{mt}\ \ \ \textgreater\ \ \ \texttt{va}  &&& 9.8 \\
            Llama 2-7B\ \  \textgreater\ \ \ \texttt{va}                                     &&& 12.2 \\
            Llama 2-7B\ \  \textgreater\ \ \ \texttt{va}\ \ \ \textgreater\ \ \ \texttt{mt}  &&& 14.4 \\
            Llama 2-7B\ \  \textgreater\ \ \ \texttt{mt}\ \ \ \textgreater\ \ \ \texttt{va}  &&& 13.8 \\
            \bottomrule
        \end{tabular}
        \vspace{-2mm}
    \end{minipage}\hfill
    \begin{minipage}{0.5\linewidth}
        \centering
        \small
        \vspace{-1mm}
        \includegraphics[width=5.7cm]{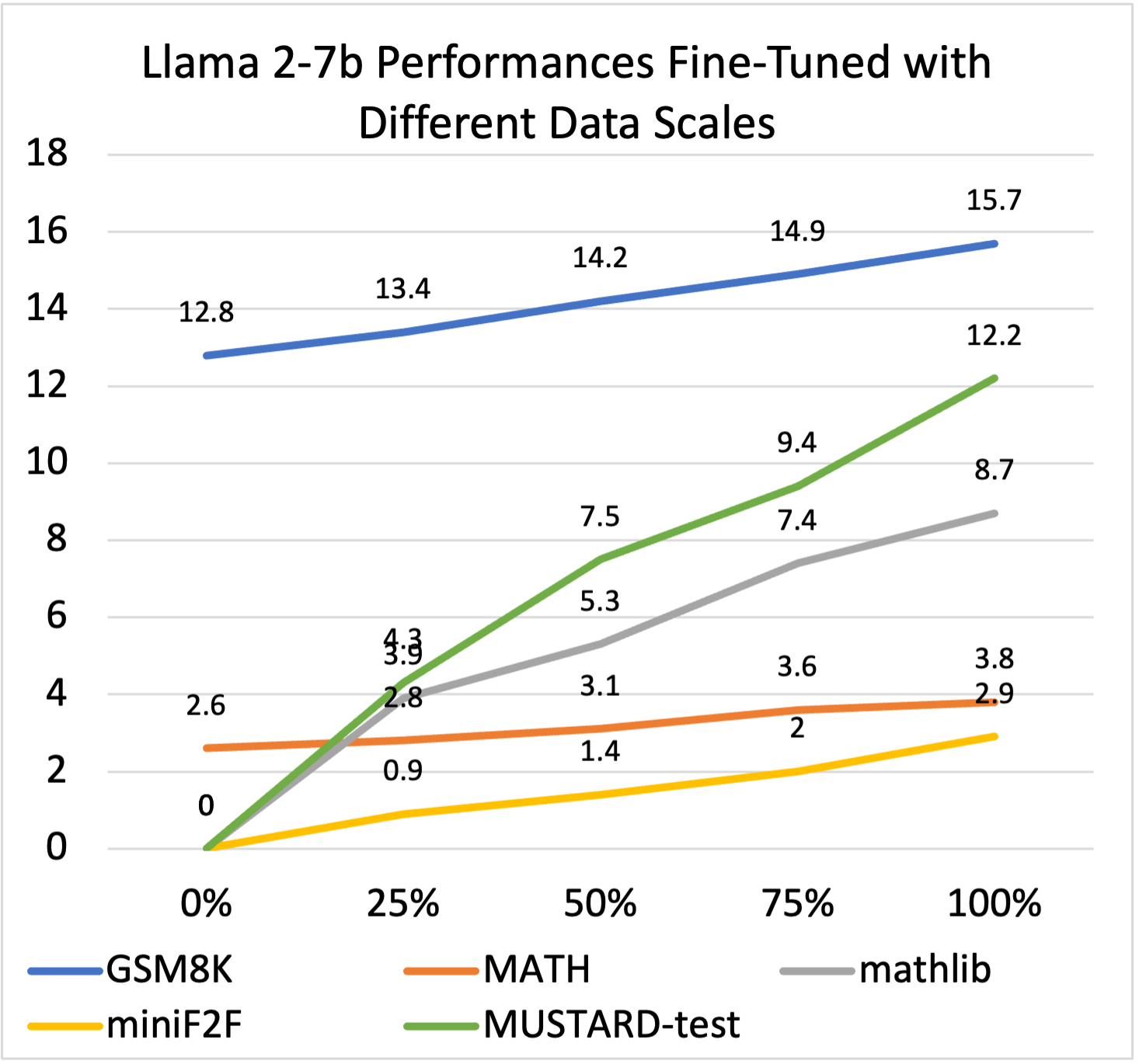}
        \vspace{-1mm}
        \captionof{figure}{
            \update{Model performances fine-tuned with different data scales. }
        }
        \vspace{-5mm}
        \label{fig:data_scale}
    \end{minipage}
    \vspace{-1mm}
\end{table}

\subsection{\update{Impact of Data Scalability}}
\vspace{-3mm}
\update{
To further study the impact of data scale on the fine-tuning results, we randomly sample 75\%, 50\%, 25\%, and 0\% data from \ourdatasetValid\ and fine-tune Llama 2-7B. 
The results are shown in Figure~\ref{fig:data_scale}. In general, the results on all datasets increase as the fine-tuning data scales up. Specifically, performances on the MUSTARD-test and mathlib have the most significant growth without a decrease in the growth rate. Therefore we expect further performance improvements when more high-quality data are included.
}

\vspace{-2.5mm}
\subsection{Pass Rate}  
\vspace{-3mm}
We study the mathematical generation ability of \ourmethod\ by investigating its pass rates on generating valid data points. 
The pass@1 results of the generated formal proofs \update{of GPT-4~\cite{DBLP:journals/corr/abs-2303-08774} and GPT-3.5~\cite{gpt35}} are shown in Table~\ref{tab:auto}.
We have the following observations. 
First of all, the overall pass@1 results are high, showing the LLMs especially GPT-4 capable of performing zero-shot mathematical reasoning. 
%
Second, the pass rates of word problems are generally higher than those of theorem proving. It indicates that the word problems are relatively easier and more familiar to the LLMs, while theorem proving is more challenging. 
Third, the all-at-once generation and step-by-step generation have similar pass rates at lower educational levels. For more challenging questions such as those at the high school level and higher educational level, step-by-step generation shows slight advantages over the all-at-once generation. This indicates that dividing and conquering (T1), (T2), and (T3) helps the model to generate higher-quality formal proofs, but the improvement is limited.
Last but not least, the improvements in pass rates after 1-step and 2-step corrections are significant. For example, theorem proving at elementary-school level with 1 seed concept improves by 22.0\% after 1-step correction, and 33.9\% after 2-step correction. Word problem at the elementary-school level with 1 seed concept improves 45.3\% after 2-step correction. The most difficult setting of generating theorem proving data at the higher-educational level with 2 seed concepts achieves 2.8\% improvement after 2-step correction, and the word problem counterpart achieves 6.1\% improvement. This indicates that the LLMs have a great potential for self-correction given the error message feedback and limited instructions. 

\begin{table}[!t]
\vspace{-4mm}
\caption{
    Pass@1 of formal proofs generated by \ourmethod. 
    k: Number of seed concepts.
}
\vspace{-2mm}
\label{tab:auto}
\scriptsize
\centering
\renewcommand{\arraystretch}{1.5} 
  \begin{tabular}{
    >{\raggedright\arraybackslash}p{0.15cm}p{0.12cm}
    rrr
    >{\raggedleft\arraybackslash}p{0.4cm}>{\raggedleft\arraybackslash}p{0.4cm}
    r
    rrr
    >{\raggedleft\arraybackslash}p{0.4cm}>{\raggedleft\arraybackslash}p{0.4cm}
  }
    \toprule
    & & \multicolumn{5}{c}{Thoerem Proving} && \multicolumn{5}{c}{Word Problem} \\    
    & & \multicolumn{3}{c}{All (\update{GPT-4})} & Step (\update{GPT-4}) & All (\update{GPT-3.5}) && \multicolumn{3}{c}{All (\update{GPT-4})} & Step (\update{GPT-4}) & All (\update{GPT-3.5}) \\
    \cmidrule{3-7} \cmidrule{9-13}
    & & \#correct=0 & 1 ($\Delta$) & 2 ($\Delta$) & 0 & 0 && 0 & 1 ($\Delta$) & 2 ($\Delta$) & 0 & 0 \\
    \midrule
    \multirow{4}*{k=1}
    	& elem   & 26.0 & 48.0 (+22.0) & 55.9 (+33.9) & 25.5 & 15.1  && 38.0 & 59.7 (+21.7) & 67.0 (+45.3) & 40.1 & 22.2 \\
    	& midd   & 16.4 & 31.8 (+15.4) & 39.6 (+24.2) & 17.0 & 4.3   && 22.9 & 39.7 (+16.8) & 47.4 (+30.6) & 28.4 & 7.0 \\
    	& high   & 6.8 & 14.4 (+7.6) & 17.2 (+9.6) & 6.6 & 1.9       && 6.7 & 16.8 (+10.1) & 21.9 (+11.8) & 8.1 & 3.4 \\
    	& higher & 2.1 & 5.6 (+3.5) & 9.8 (+6.3) & 3.0 & 0.7         && 3.6 & 10.9 (+7.3) & 16.0 (+8.7) & 4.3 & 2.7 \\
    \midrule
    \multirow{4}*{k=2}
    	& elem   & 24.1 & 42.3 (+18.2) & 52.3 (+34.1) & 23.2 & 12.8 && 32.2 & 49.5 (+17.3) & 58.2 (+40.9) & 31.9 & 22.1 \\
    	& midd   & 14.0 & 25.2 (+11.2) & 34.1 (+22.9) & 15.0 & 5.3 && 16.9 & 27.3 (+10.4) & 34.5 (+24.1) & 17.3 & 5.7 \\
    	& high   & 3.8 & 8.3 (+4.5) & 12.1 (+7.6) & 5.4 & 2.2      && 5.7 & 11.0 (+5.3) & 16.2 (+10.9) & 4.5 & 2.5 \\
    	& higher & 1.1 & 3.3 (+2.2) & 5.0 (+2.8) & 2.6 & 1.3       && 2.6 & 7.0 (+4.4) & 10.5 (+6.1) & 2.9 & 2.1 \\
    \bottomrule
  \end{tabular}
  \vspace{-2mm}
\end{table}

\vspace{-2mm}
\subsection{Diversity and Difficulty}
\begin{figure}[!t]
    \centering
    \vspace{-2mm}
    \includegraphics[width=\textwidth]{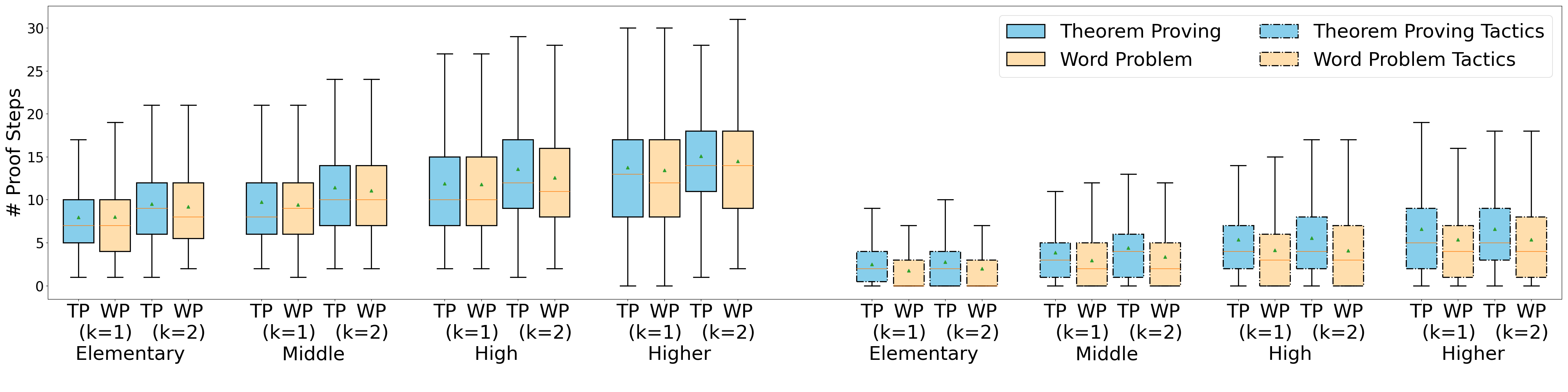}
    \caption{Distributions of formal proof lengths.}
    \vspace{-3mm}
    \label{fig:proof_steps}
\end{figure}

\vspace{-3mm}
We compute ROUGE-L \citep{lin-2004-rouge} to check the diversity of generated informal statements and proofs. 
The resulting ROUGE-L scores are below 0.25 and indicate high data diversity. 
We demonstrate detailed computation and results in Appendix~\ref{app:diversity}.
%
We then investigate the proof lengths in \ourdatasetinit\ and the distributions are demonstrated in Figure~\ref{fig:proof_steps}. 
We count both reasoning steps of formal statement-proof pairs and steps of formal proof only, which are shown on the left- and right-hand side of Figure~\ref{fig:proof_steps}, respectively.
It is demonstrated that proof length increases over educational levels.
Solving elementary problems needs about 5 to 10 steps while
solving higher-educational problems requires a median number between 10 to 15 steps.
The most challenging problems require around 30 reasoning steps or about 20 formal proof steps.
Therefore, \ourdataset\ produces diverse mathematical problems with multiple topics and difficulties.

\vspace{-2mm}
\section{Conclusion}
\label{sec:conclusion}
\vspace{-3mm}
In this paper, we introduce \ourmethod\ to automatically generate mathematical datasets with high-quality solutions that cover a variety of mathematical skills. 
Leveraging the LLM and Lean Prover, \ourmethod\ can generate the problem statement, informal solution, and formal solution. and use a Lean Prover to automatically verify the formal solution and provide feedback for revision. 
At last, we apply the proposed \ourmethod\ and obtain 5,866 problems with step-by-step solutions that cover different educational levels and mathematical abilities. The obtained dataset has shown its high quality, diversity, and effectiveness in improving language models' mathematical reasoning performance, showing the great potential of our proposed \ourmethod\ and dataset in further research on language models.

\section*{Acknowledgements}
This work was supported in part by the 
National Key R\&D Program of China under Grant No. 2020AAA0109700, 
Guangdong Outstanding Youth Fund (Grant No. 2021B1515020061), 
National Natural Science Foundation of China (NSFC) under Grant No.61976233, 
Mobility Grant Award under Grant No. M-0461, 
Shenzhen Science and Technology Program (Grant No. GJHZ20220913142600001), 
Nansha Key RD Program under Grant No.2022ZD014,
National Natural Science Foundation of China under Grant No.62006255,
and the National Natural Science Foundation of China under Grant 62371411, the Research Grants Council of the Hong Kong SAR under Grant GRF 11217823, InnoHK initiative, the Government of the HKSAR, Laboratory for AI-Powered Financial Technologies.
We thank MindSpore for the partial support of this work, which is a new deep learning computing framework\footnote{https://www.mindspore.cn/}.

The authors would also like to thank Hui Jin, Jianhao Shen, Chengwu Liu, Cen Li, and Junhao Cheng for their hard work on the manual check related to Section~\ref{sec:human_eval}.



\bibliography{iclr2024_ATP_references}
\bibliographystyle{iclr2024_conference}

\appendix

\section{\update{Future Works}}
\update{
Current formal validation still suffers mild inconsistency between the \#reduce statements and various kinds of theorem proofs. 
In the future, we will explore more rigorous and careful data filtering.
We will also explore data generation and mathematical reasoning via the same language model, which is an interesting setup to study large language models' proficiency of mathematical reasoning.
Moreover, the ablation study on data scalability shows consistent performance increases when more data from MUSTARD are introduced, suggesting a great potential for scalability. Fortunately, MUSTARD reduces the cost of acquiring such high-quality step-by-step complex reasoning data and obtains correct, scalable, and reusable data. Therefore, in future work, we would love to build a community in which all members can join the data synthesis process, and acquire and share more high-quality data with the whole community.
}

\section{Mathematical Concepts}  
\label{app:concepts}
Table~\ref{tab:concept} shows the concept statistics. 
Concepts are grouped into multiple domains mainly according to subjects, except that at the elementary level are grouped by grades due to the lack of subject division. 
Tables~\ref{tab:app_concepts_elem}, \ref{tab:app_concepts_midd}, \ref{tab:app_concepts_high} and \ref{tab:app_concepts_higher} demonstrate the detailed concepts in each domain.

\begin{table}[!h] 
    \caption{Statistics of the mathematical concept pool. }
    \label{tab:concept}
    \small
    \centering
\begin{tabular}{
        c >{\centering\arraybackslash}p{0.05cm} c
        c >{\centering\arraybackslash}p{0.05cm} c
        c >{\centering\arraybackslash}p{0.05cm} c
        c >{\centering\arraybackslash}p{0.05cm}
    }
    \toprule
    \multicolumn{2}{c}{Elementary School} && \multicolumn{2}{c}{Middle School} && \multicolumn{2}{c}{High School} && \multicolumn{2}{c}{Higher Education} \\
    \cmidrule{1-2}  \cmidrule{4-5}  \cmidrule{7-8}  \cmidrule{10-11} 
    Domains & \# && Domains & \# && Domains & \# && Domains & \# \\
\bottomrule
    \end{tabular}
\end{table}

\begin{table}[!h]  
    \caption{Mathematical concepts at the level of elementary school. }
    \label{tab:app_concepts_elem}
    \begin{tabular}{
        >{\centering\arraybackslash}p{.2\textwidth}l
    }
    \toprule
    \textbf{Domain} & \textbf{Concept} \\
    \midrule
        1st grade & \makecell[{{>{\raggedright\arraybackslash}p{.7\textwidth}}}]{
            ``Place value'',
            ``Addition and subtraction'',
            ``Measurement, data, and geometry'',
    } \\ \cmidrule{2-2}
        2nd grade & \makecell[{{>{\raggedright\arraybackslash}p{.7\textwidth}}}]{
            ``Add and subtract within 20'',
            ``Place value'',
            ``Add and subtract within 100'',
            ``Add and subtract within 1,000'',
            ``Money and time'',
            ``Measurement'',
            ``Data'',
            ``Geometry'',
    } \\ \cmidrule{2-2}
        3rd grade & \makecell[{{>{\raggedright\arraybackslash}p{.7\textwidth}}}]{
            ``Intro to multiplication'',
            ``1-digit multiplication'',
            ``Addition, subtraction, and estimation'',
            ``Intro to division'',
            ``Understand fractions'',
            ``Equivalent fractions and comparing fractions'',
            ``More with multiplication and division'',
            ``Arithmetic patterns and problem solving'',
            ``Quadrilaterals'',
            ``Area'',
            ``Perimeter'',
            ``Time'',
            ``Measurement'',
            ``Represent and interpret data'',
    } \\
\bottomrule    
    \end{tabular}
\end{table}

\begin{table}[!h]  
    \caption{Mathematical concepts at the level of middle school. }
    \label{tab:app_concepts_midd}
    \begin{tabular}{
        cl
    }
    \toprule
    \textbf{Domain} & \textbf{Concept} \\
    \midrule
        7th grade & \makecell[{{>{\raggedright\arraybackslash}p{.7\textwidth}}}]{
            ``Negative numbers: addition and subtraction'',
            ``Negative numbers: multiplication and division'',
            ``Fractions, decimals, \& percentages'',
            ``Rates \& proportional relationships'',
            ``Expressions, equations, \& inequalities'',
            ``Geometry'',
            ``Statistics and probability'',
    } \\ \cmidrule{2-2}
        8th grade & \makecell[{{>{\raggedright\arraybackslash}p{.7\textwidth}}}]{
            ``Numbers and operations'',
            ``Solving equations with one unknown'',
            ``Linear equations and functions'',
            ``Systems of equations'',
            ``Geometry'',
            ``Geometric transformations'',
            ``Data and modeling'',
    } \\ \cmidrule{2-2}
        Algebra basics & \makecell[{{>{\raggedright\arraybackslash}p{.7\textwidth}}}]{
            ``Foundations'',
            ``Algebraic expressions'',
            ``Linear equations and inequalities'',
            ``Graphing lines and slope'',
            ``Systems of equations'',
            ``Expressions with exponents'',
            ``Quadratics and polynomials'',
            ``Equations and geometry'',
    } \\
\bottomrule    
    \end{tabular}
\end{table}

\begin{table}[!h]  
    \caption{Mathematical concepts at the level of high school. }
    \label{tab:app_concepts_high}
    \begin{tabular}{
        cl
    }
    \toprule
    \textbf{Domain} & \textbf{Concept} \\
    \midrule
        Algebra 1 & \makecell[{{>{\raggedright\arraybackslash}p{.7\textwidth}}}]{
            ``Algebra foundations'',
            ``Solving equations \& inequalities'',
            ``Working with units'',
            ``Linear equations \& graphs'',
            ``Forms of linear equations'',
            ``Systems of equations'',
            ``Inequalities (systems \& graphs)'',
            ``Functions'',
            ``Sequences'',
            ``Absolute value \& piecewise functions'',
            ``Exponents \& radicals'',
            ``Exponential growth \& decay'',
            ``Quadratics: Multiplying \& factoring'',
            ``Quadratic functions \& equations'',
            ``Irrational numbers'',
            ``Creativity in algebra'',
    } \\ \cmidrule{2-2}   
        Algebra 2 & \makecell[{{>{\raggedright\arraybackslash}p{.7\textwidth}}}]{
            ``Polynomial arithmetic'',
            ``Complex numbers'',
            ``Polynomial factorization'',
            ``Polynomial division'',
            ``Polynomial graphs'',
            ``Rational exponents and radicals'',
            ``Exponential models'',
            ``Logarithms'',
            ``Transformations of functions'',
            ``Equations'',
            ``Trigonometry'',
            ``Modeling'',
    } \\ \cmidrule{2-2}   
        High school geometry & \makecell[{{>{\raggedright\arraybackslash}p{.7\textwidth}}}]{
            ``Performing transformations'',
            ``Transformation properties and proofs'',
            ``Congruence'',
            ``Similarity'',
            ``Right triangles \& trigonometry'',
            ``Analytic geometry'',
            ``Conic sections'',
            ``Circles'',
            ``Solid geometry'',
    } \\
\bottomrule    
    \end{tabular}
\end{table}

\begin{table}[!h]  
    \caption{Mathematical concepts at the level of higher education. }
    \label{tab:app_concepts_higher}
    \begin{tabular}{
        cl
    }
    \toprule
    \textbf{Domain} & \textbf{Concept} \\
    \midrule
        AP College Statistics & \makecell[{{>{\raggedright\arraybackslash}p{.7\textwidth}}}]{
            ``Exploring categorical data'',
            ``Exploring one-variable quantitative data: Displaying and describing'',
            ``Exploring one-variable quantitative data: Summary statistics'',
            ``Exploring one-variable quantitative data: Percentiles, z-scores, and the normal distribution'',
            ``Exploring two-variable quantitative data'',
            ``Collecting data'',
            ``Probability'',
            ``Random variables and probability distributions'',
            ``Sampling distributions'',
            ``Inference for categorical data: Proportions'',
            ``Inference for quantitative data: Means'',
            ``Inference for categorical data: Chi-square'',
            ``Inference for quantitative data: slopes'',
            ``Prepare for the 2022 AP Statistics Exam'',
    } \\ \cmidrule{2-2}   
        College Algebra & \makecell[{{>{\raggedright\arraybackslash}p{.7\textwidth}}}]{
            ``Linear equations and inequalities'',
            ``Graphs and forms of linear equations'',
            ``Functions'',
            ``Quadratics: Multiplying and factoring'',
            ``Quadratic functions and equations'',
            ``Complex numbers'',
            ``Exponents and radicals'',
            ``Rational expressions and equations'',
            ``Relating algebra and geometry'',
            ``Polynomial arithmetic'',
            ``Advanced function types'',
            ``Transformations of functions'',
            ``Rational exponents and radicals'',
            ``Logarithms'',
    } \\
\bottomrule    
    \end{tabular}
\end{table}

\newpage
\section{Prompt Templates in \ourmethod}  
\subsection{Prompt Template for Proof Filtering}  
\label{app:proof_filtering}
Table~\ref{tab:prompt_proof_filtering} demonstrates the prompt template used in the proof-filtering stage in \ourmethod.

\begin{table}[!t]
    \caption{Prompt template for proof filtering.}
    \label{tab:prompt_proof_filtering}
    \small
    \begin{tabular}{p{\textwidth}}
        \toprule
        \makecell[l]{\textbf{Prompt Template for Proof Filtering}} \\
        \midrule
         \texttt{In the following, you are given a ``Problem'', a pair of corresponding ``Informal proof'' and ``Formal proof in Lean 3'', along with error messages from a Lean Prover corresponding to the ``Formal proof in Lean 3''. Now please carefully modify the ``Formal proof in Lean 3'' section so that it passes the Lean Prover without error. You should write the modified complete proof in your response.} \\ \\

        \texttt{\# Problem: <generated problem>} \\ \\

        \texttt{\# Informal proof: <generated informal proof>} \\ \\ 

        \texttt{\# Formal proof (1) in Lean 3: } \\ 
        \texttt{```lean} \\ 
        \texttt{line 1 <code> } \\ 
        \texttt{line 2 <code> } \\ 
        \texttt{line 3 <code> } \\ 
        \texttt{... } \\ 
        \texttt{```} \\ \\

        \texttt{\# Error messages for Formal proof (1) from Lean Prover: } \\ 
        \texttt{<error messages>}\\ \\

        \texttt{...} \\ \\

        \texttt{\# Formal proof (k) in Lean 3: } \\ 
        \texttt{```lean} \\ 
        \texttt{line 1 <code> } \\ 
        \texttt{line 2 <code> } \\ 
        \texttt{line 3 <code> } \\ 
        \texttt{... } \\ 
        \texttt{```} \\ \\

        \texttt{\# Error messages for Formal proof (k) from Lean Prover: } \\ 
        \texttt{<error messages>}\\
        \bottomrule
    \end{tabular}
\end{table}

\begin{table}[!t]
    \vspace{-1mm}
    \caption{Prompt templates for step-by-step generation in the proof-generation stage.}
    \label{tab:prompt_step_by_step}
    \small
    \vspace{-1mm}
    \begin{tabular}{
        p{.04\textwidth}p{.92\textwidth}
        }
        \toprule
        \makecell[c]{\textbf{Prompt Templates for Step-by-Step Generation}} \\
        \hline
        {(T1)} & 
        \makecell[{{>{\raggedright\arraybackslash}p{.92\textwidth}}}]{
            \texttt{You are a math expert. Now please come up with a math problem according to the following requirements. The math problem should contain a question part (indicated by ``Problem: ''), a corresponding solution in natural language (indicated by ``Informal proof:''), and a translated formal solution in Lean 3 (indicated by ``Formal proof in Lean 3:''). Please note that the informal proof and the formal proof need to be identical. Please create a [QUESTION TYPE] in the level of [EDUCATIONAL LEVEL] based on the following knowledge point(s): \*[CONCEPT]\* in [DOMAIN]; \*[CONCEPT]\* in [DOMAIN].}\\ \\
            \texttt{Please first write the question part regardless of the other parts. You must write the following format, filling in the ``\# Problem: '' section, and leaving the other two sections empty.} \\ \\
            \texttt{\# Problem: ...} \\ \\         
            \texttt{\# Informal proof: ...} \\ \\        
            \texttt{\# Formal proof in Lean 3: ...} \\ 
        } \\
\bottomrule
    \end{tabular}
    \vspace{-1mm}
\end{table}

\subsection{Prompt Templates for Step-by-Step Generation}  
\label{app:stepbystep}
Table~\ref{tab:prompt_step_by_step} demonstrates the variation of prompt templates used in the proof-generation stage in \ourmethod. 
In this variation, an LLM is conducted to perform (T1), (T2), and (T3) separately to generate the informal statement, informal solution, and formal solution.  
It is noted that to prompt the LLM to fulfill (T3), we assign the character of ``a master in Lean'' rather than the previous ``a math expert'' to obtain higher quality Lean proofs.



\newpage
\section{More Statistic Results of \ourdataset}
\subsection{Difficulty of \ourdataset\ by Number of Correction}
\begin{figure}[!h]
    \centering
    \includegraphics[width=.7\textwidth]{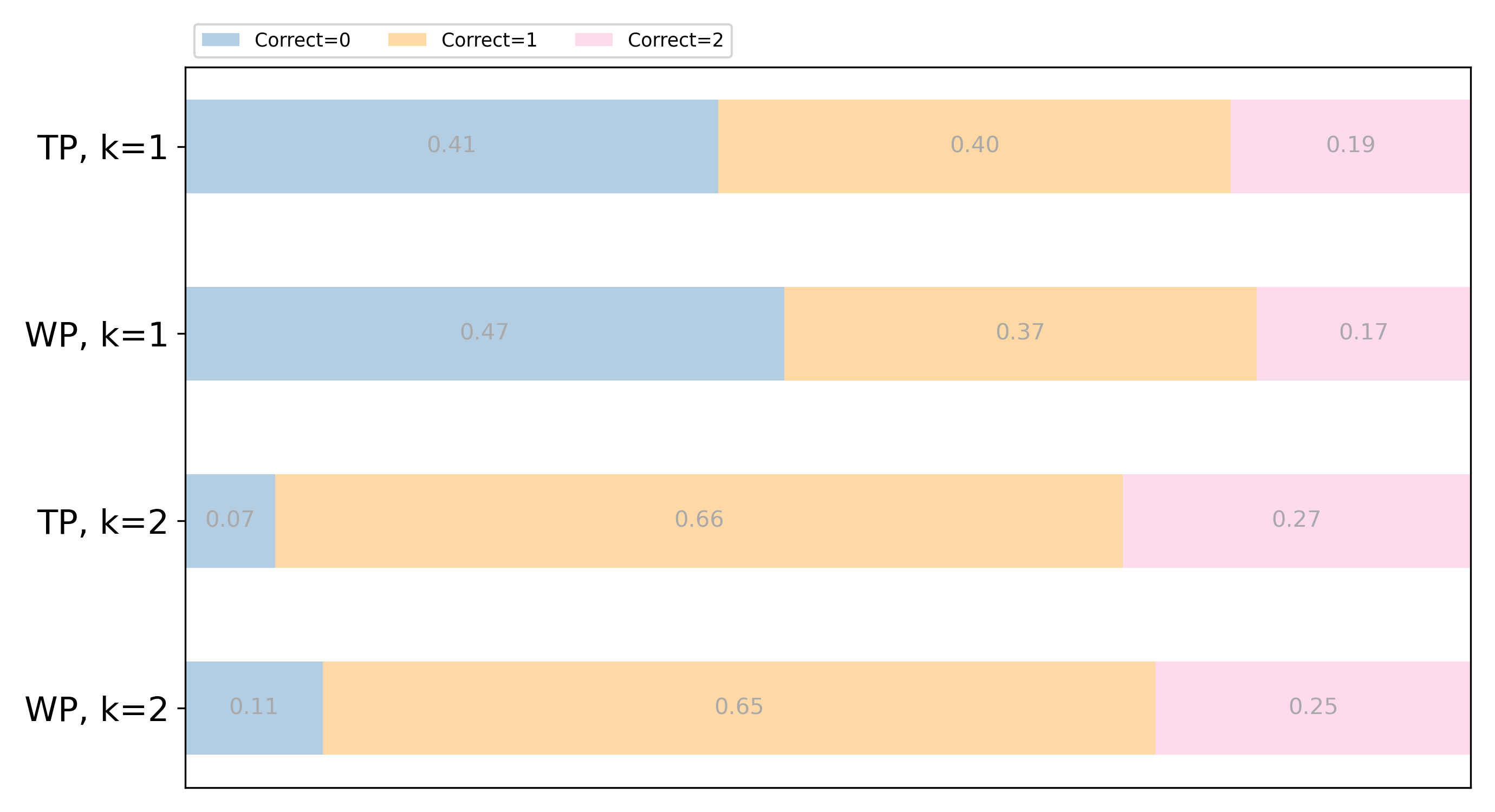}
    \caption{Data proportions according to correction steps.}
    \label{fig:correction}
\end{figure}

Figure~\ref{fig:correction} demonstrates the proportions of data points that obtain valid proof after different numbers of corrections. 
Generally speaking, data points without correction are relatively less difficult for the LLMs, while those that require multiple corrections are challenging. 
Overall, theorem-proving problems are more challenging for LLMs to solve than the generated word problems. 
For data points with 2 seed concepts, more than 90\% of the data can not pass the prover validation at the first generation. 
And almost 30\% of them cost 2 correction steps to obtain valid proof. 
Similar observations are found in word problems with 2 seed concepts. 
It is suggested that the data subset with 2 seed concepts is challenging to the LLMs in general. 
In contrast, data with 1 seed concept are easier for LLMs. But there are still more than half data points that need proof improvements based on error messages from the theorem prover. 
Therefore, overall, \ourmethod\ generates valid data points in different difficulty levels, and the majority of the problems are challenging for the LLMs. 

\subsection{Data Diversity}
\label{app:diversity}
\begin{figure*}[!t]
    \centering
    \begin{subfigure}[b]{.49\textwidth}
        \centering
        \includegraphics[width=\textwidth]{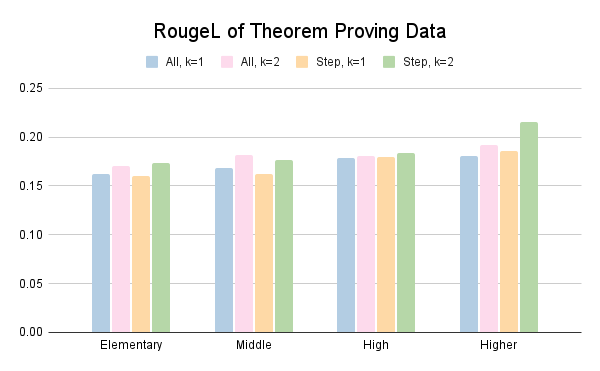}
    \end{subfigure}
    \begin{subfigure}[b]{.49\textwidth}  
        \centering 
        \includegraphics[width=\textwidth]{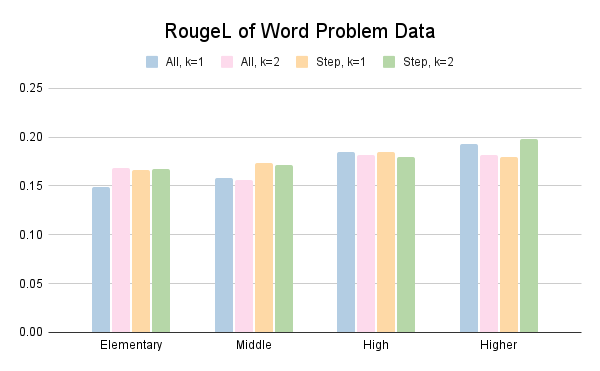}
    \end{subfigure}
    \caption{
    Diversity of informal statements and proofs by ROUGE.
    All: All-at-once generation. 
    Step: Step-by-step generation.
    k: Number of seed concepts.
    } 
    \label{fig:rouge}
\end{figure*}

We compute ROUGE-L \citep{lin-2004-rouge} to check the diversity of generated informal statements and proofs. 
Specifically, given a data set, we perform 10 rounds of bootstrapping.
In each round, we randomly sample 10 data points from the data set, each of which is paired with the remaining data points, and compute pair-wise ROUGE-L scores. 
The ROUGE-L score per round is obtained by averaging the pair-wise scores. 
The final ROUGE-L score is an average score over the bootstrapping. 
We compare the scores among the generation settings, and the results are shown in Figure~\ref{fig:rouge}.

The results show that all settings have a ROUGE-L score beneath 0.25, which indicates a high diversity of the generated informal statements and informal proofs. 
All-at-once and step-by-step generation share similar data diversity. 
The ROUGE-L scores slightly increase over the educational levels. 
Generating higher-educational theorem proving data with 2 concepts step-by-step exceeds 0.2 in ROUGE-L.
Therefore, this setting is relatively challenging.

\begin{figure*}[!h]
    \centering
    \begin{subfigure}[b]{.47\textwidth}
        \centering
        \includegraphics[width=.9\textwidth]{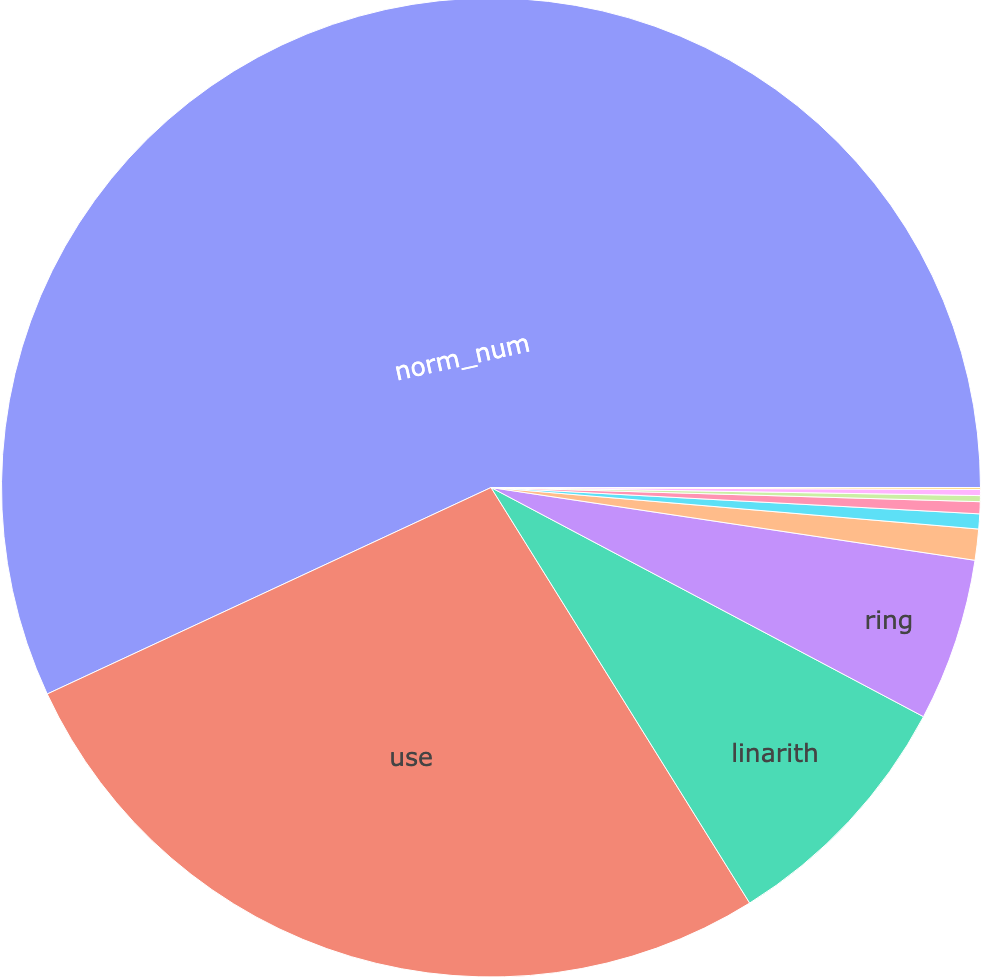}
    \end{subfigure}    
    \hfill
    \begin{subfigure}[b]{.47\textwidth}
        \centering
        \includegraphics[width=.9\textwidth]{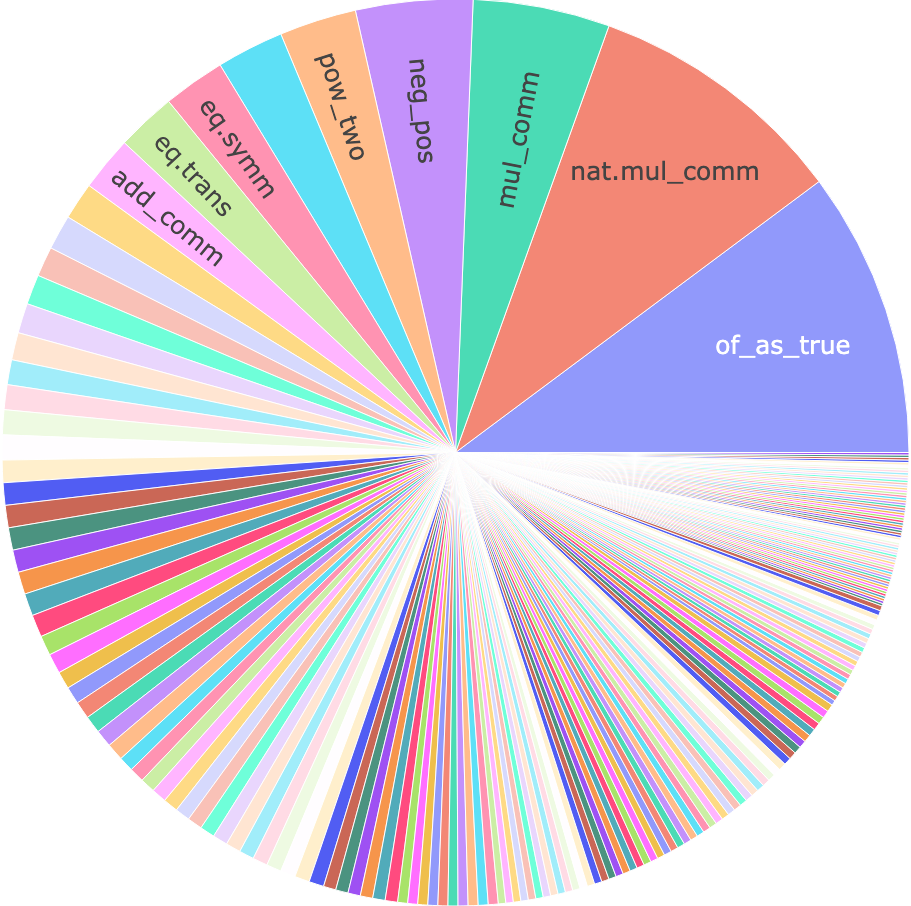}
    \end{subfigure}
    \hfill 
    \hfill 
    \hfill
    \begin{subfigure}[b]{.47\textwidth}  
        \centering 
        \includegraphics[width=.9\textwidth]{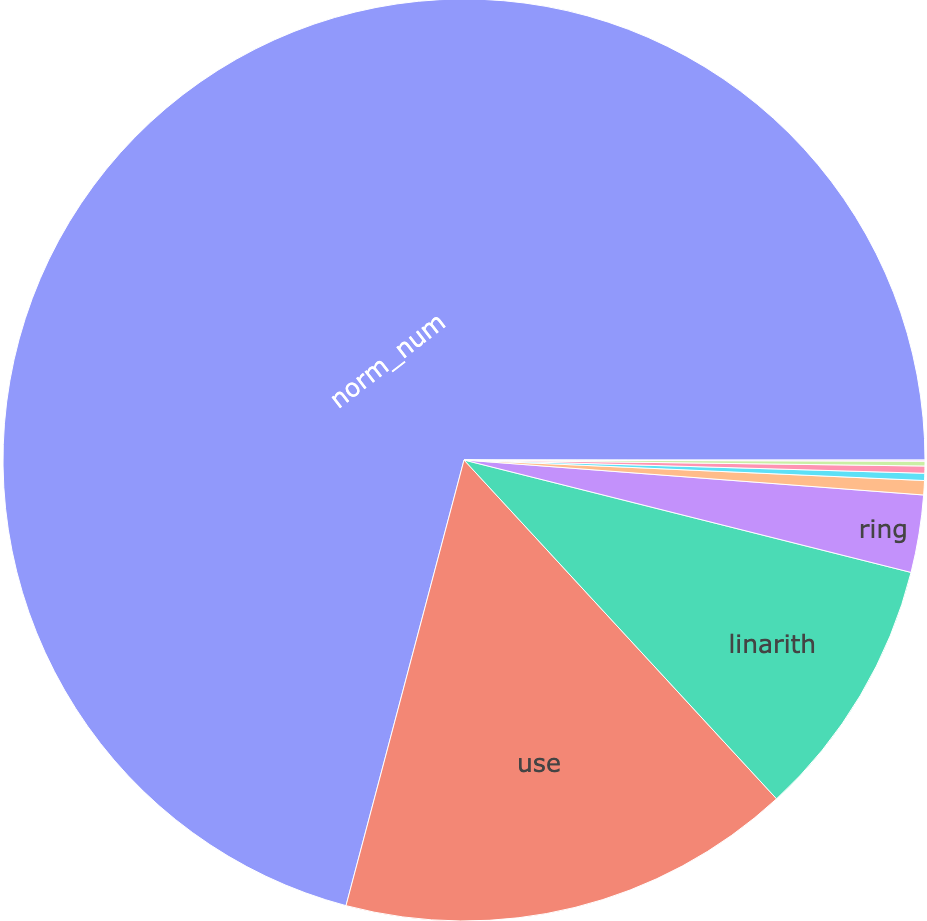}
    \end{subfigure}
    \hfill
    \begin{subfigure}[b]{.47\textwidth}  
        \centering 
        \includegraphics[width=.9\textwidth]{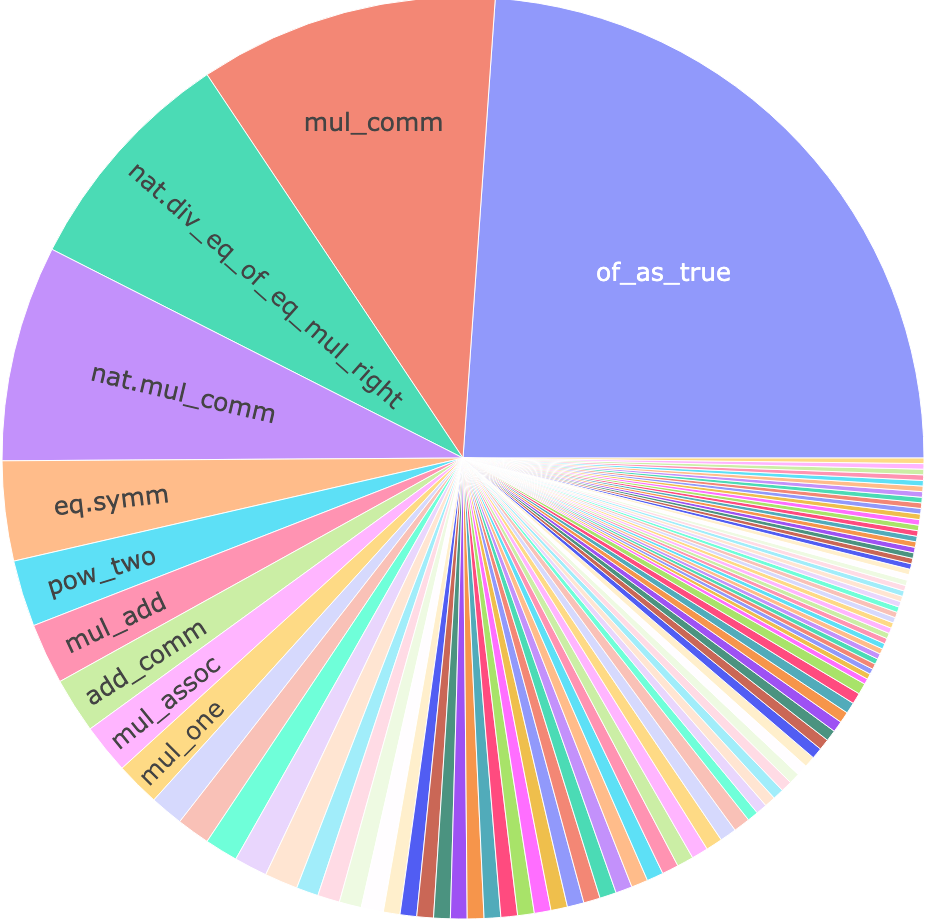}
    \end{subfigure}
    \caption{
    Distribution of tactics (left-hand side) and lemmas (right-hand side) used by the formal statements and proofs in theorem proving (upper row) and word problem (lower row).
    } 
    \label{fig:tactic}
\end{figure*}

We further investigate the diversity of formal statements and proofs.
We collect all the occurrences of used tactics and lemmas, and Figure~\ref{fig:tactic} shows the distributions. 
It is demonstrated that LLMs tend to use diverse lemmas to solve the problems. 
And LLMs memorize and understand multiple lemmas. 
The most commonly used lemmas in both theorem proving and word problems include \texttt{of\_as\_true}, \texttt{mul\_comm}, and \texttt{nat.mul\_comm}.
\texttt{of\_at\_true} is often used in automated proofs for complex propositions to first prove their decidability. 
\texttt{mul\_comm} and \texttt{nat.mul\_comm} are about the multiplicative commutative law and are often used for expression simplification.
Similarly, the frequently used \texttt{neg\_pos}, \texttt{pow\_two}, \texttt{nat.div\_eq\_of\_eq\_mul\_right} and so forth are basic lemmas for proving advanced properties. 
{Similar observations are found in tactics.}
Therefore, LLMs are good at breaking problems into basic steps and using basic lemmas to complete complex proving.

%



\newpage
\section{Case Study}
\label{app:case_study}

Tables~\ref{tab:good_case_1} to \ref{tab:bad_case_7} demonstrates successful and failed cases generated by \ourmethod.

\begin{table}[!h]
    \caption{A successful case generated by \ourmethod.}
    \label{tab:good_case_1}
    \small
    \centering
    \begin{tabular}{
        p{\textwidth}
    }
        \toprule
        \textbf{Question Type: Word Problem.\quad Educational Level: Elementary School.\quad k=1. \quad\quad\Green{Successful case. }} \\
        \midrule
        \textbf{Concept(s): } \textit{Division} in 4th grade. \\
        \midrule
        \textbf{Comments: } 
        \ourmethod\ accurately counts the number of people and correctly performs division.
         \\
        \midrule
        \textbf{Informal Statement} \\
        {John and his 3 friends are on a trip. They bought a huge box of chocolates containing 240 pieces. They decided to split the chocolates equally among themselves. How many chocolates will each one of them get?} \\ \\
         
        \textbf{Informal Proof} \\
        {Since there are 4 people in total and they have 240 chocolates, we can find out how many chocolates each person will get by dividing the total number of chocolates by the total number of people. So, 240 divided by 4 equals 60. Therefore, each person will get 60 chocolates.} \\ \\
         
        \textbf{Formal Statement and Proof} \\
        \texttt{\textbf{\textcolor{purple}{def}} divide (x y : $\mathbb{N})$ : $\mathbb{N}$ := x / y} \\ \\
        \texttt{\#reduce divide 240 4  \textcolor{gray}{\textit{-- The output should be 60, which means each person will get 60 chocolates. }}} \\
        \bottomrule
    \end{tabular}
\end{table}

\begin{table}[!h]
    \caption{A failed cased generated by \ourmethod.}
    \label{tab:bad_case_1}
    \small
    \centering
    \begin{tabular}{
        p{\textwidth}
    }
        \toprule
        \textbf{Question Type: Theorem Proving.\quad Educational Level: Elementary School.\quad k=1. \quad\quad \Red{Failed case. } } \\
        \midrule
        \textbf{Concept(s): } \textit{Quadrilaterals} in 3rd grade. \\
        \midrule
        \textbf{Comments: } 
        \ourmethod\ believes that a quadrilateral with four equal sides is a square. It treats a necessary condition as a sufficient condition.\\
        \midrule
        \textbf{Informal Statement} \\
        Given a quadrilateral ABCD where AB = BC = CD = DA = 5 units, prove that this quadrilateral is a square. \\ \\

        \textbf{Informal Proof} \\
        The given quadrilateral ABCD has all its sides equal. This is a property of a square. Hence, ABCD is a square. \\ \\

        \textbf{Formal Statement and Proof} \\
        \texttt{\textbf{\textcolor{purple}{import}} data.real.basic} \\ \\

        \texttt{\textbf{\textcolor{purple}{def}} is\_square\_quad (AB BC CD DA : $\mathbb{R}$) : Prop := AB = BC $\land$ BC = CD $\land$ CD = DA} \\ \\
        
        \texttt{\textbf{\textcolor{purple}{def}} ABCD\_is\_square : $\forall$ (AB BC CD DA : $\mathbb{R}$), AB = 5 $\land$ BC = 5 $\land$ CD = 5 $\land$ DA = 5 $\to$ is\_square\_quad AB BC CD DA :=} \\
        \texttt{\textbf{\textcolor{purple}{begin}}} \\
        \texttt{   \tacticBlue{intros} AB BC CD DA h,} \\
        \texttt{   \tacticBlue{cases} h \textbf{\textcolor{purple}{with}} h$_1$ h$_2$,} \\
        \texttt{   \tacticBlue{cases} h$_2$ \textbf{\textcolor{purple}{with}} h$_3$ h$_4$,} \\
        \texttt{   \tacticBlue{cases} h$_4$ \textbf{\textcolor{purple}{with}} h$_5$ h$_6$,} \\
        \texttt{   \tacticBlue{split},} \\
        \texttt{   \{ \tacticBlue{rw} h$_1$, \tacticBlue{exact} h$_3$ \},} \\
        \texttt{   \tacticBlue{split},} \\
        \texttt{   \{ \tacticBlue{rw} h$_3$, \tacticBlue{exact} h$_5$ \},} \\
        \texttt{   \{ \tacticBlue{rw} h$_5$, \tacticBlue{exact} h$_6$ \},} \\
        \texttt{\textbf{\textcolor{purple}{end}} } \\
        \bottomrule
    \end{tabular}
\end{table}

\begin{table}[!h]
    \caption{A successful case generated by \ourmethod.}
    \label{tab:good_case_2}
    \small
    \centering
    \begin{tabular}{
        p{\textwidth}
    }
        \toprule
        \textbf{Question Type: Word Problem.\quad Educational Level: Elementary School.\quad k=2. \quad\quad \Green{Successful case. }} \\
        \midrule
        \textbf{Concept(s): } \textit{Addition and subtraction} in 1st grade; Measuring angles in 4th grade. \\
        \midrule
        \textbf{Comments: } 
        \ourmethod\ creatively combines the two concepts and constructs a reasonable question. \\
        \midrule
        \textbf{Informal Statement} \\
        John has a favorite toy car that he likes to move around a circular track. He starts moving the car from a certain point and stops after the car has moved 90 degrees. Then he moves the car again but this time in the opposite direction, stopping when the car has moved 45 degrees. What is the net degree movement of John's toy car? \\ \\
        \textbf{Informal Proof} \\
        Initially, John moves his car 90 degrees. Then he moves the car in the opposite direction 45 degrees. Since the car is moving in the opposite direction, we consider this as -45 degrees. To find the net movement, we add the two movements together. So, 90 + (-45) equals 45. Therefore, the net degree movement of John's toy car is 45 degrees. \\ \\
        \textbf{Formal Statement and Proof} \\
        \texttt{\textbf{\textcolor{purple}{import}} data.int.basic} \\ \\

		\texttt{\textbf{\textcolor{purple}{open}} int} \\ \\

		\texttt{\textit{\textcolor{gray}{-- define the initial movement}}} \\
		\texttt{\textbf{\textcolor{purple}{def}} initial\_movement : $\mathbb{Z}$ := 90} \\ \\

		\texttt{\textit{\textcolor{gray}{-- define the backward movement}}} \\
		\texttt{\textbf{\textcolor{purple}{def}} backward\_movement : $\mathbb{Z}$ := -45} \\ \\

		\texttt{\textit{\textcolor{gray}{-- define the function to calculate the net movement}}} \\
		\texttt{\textbf{\textcolor{purple}{def}} net\_movement (initial backward : $\mathbb{Z}$) : $\mathbb{Z}$ := initial + backward} \\ \\

		\texttt{\textit{\textcolor{gray}{-- proof that the net movement is 45 degrees}}} \\
		\texttt{\textbf{\textcolor{purple}{example}} : net\_movement initial\_movement backward\_movement = 45 :=} \\ 
		\texttt{\textbf{\textcolor{purple}{begin}}} \\
		\texttt{   \tacticBlue{unfold} net\_movement,} \\
		\texttt{   \tacticBlue{unfold} initial\_movement,} \\
		\texttt{   \tacticBlue{unfold} backward\_movement,} \\
		\texttt{   \tacticBlue{exact} dec\_trivial,} \\
		\texttt{\textbf{\textcolor{purple}{end}}} \\
        \bottomrule
    \end{tabular}
\end{table}

\begin{table}[!h]
    \caption{A successful case generated by \ourmethod.}
    \label{tab:good_case_3}
    \small
    \centering
    \begin{tabular}{
        p{\textwidth}
    }
        \toprule
        \textbf{Question Type: Theorem Proving.\quad Educational Level: Middle School.\quad k=2. \quad\quad \Green{Successful case. }} \\
        \midrule
        \textbf{Concept(s): } \textit{Geometry} in 8th grade; \textit{Algebraic expressions} in Algebra basics. \\
        \midrule
        \textbf{Comments: } 
        \ourmethod\ understands polynomials and writes a valid proof in Lean. \\
        \midrule
        \textbf{Informal Statement} \\
        Given a rectangle ABCD where AB is x + 5 and AD is 2x - 3. Prove that the area of the rectangle ABCD is $(2x^2 + 7x - 15)$ square units. \\ \\
        \textbf{Informal Proof} \\
        In a rectangle, the area is calculated by multiplying the length and the width. Here, the length is represented by the expression $x + 5$ and the width is represented by the expression $2x - 3$. By multiplying these two expressions together, we get: \\ \\
        
        $(x + 5) * (2x - 3) = 2x^2 - 3x + 10x - 15 = 2x^2 + 7x - 15$ \\ \\

        Therefore, the area of the rectangle ABCD is $(2x^2 + 7x - 15)$ square units. \\ \\

        \textbf{Formal Statement and Proof} \\
        \texttt{\textbf{\textcolor{purple}{import}} algebra.ring.basic} \\ \\

		\texttt{\textbf{\textcolor{purple}{variables}} (x : $\mathbb{Z}$)} \\ \\

		\texttt{\textbf{\textcolor{purple}{def}} rectangle\_area : (x + 5) * (2 * x - 3) = 2 * (x * x) + 7 * x - 15 :=} \\
		\texttt{\textbf{\textcolor{purple}{begin}}} \\
		\texttt{   \textbf{\textcolor{purple}{have}} h$_1$ : (x + 5) * (2 * x - 3) = 2 * (x * x) - 3 * x + 10 * x - 15, \textbf{\textcolor{purple}{by}} ring,} \\
		\texttt{   \textbf{\textcolor{purple}{have}} h$_2$ : 2 * (x * x) - 3 * x + 10 * x - 15 = 2 * (x * x) + 7 * x - 15, \textbf{\textcolor{purple}{by}} \tacticBlue{simp},} \\
		\texttt{   \tacticBlue{rw} h$_1$,} \\
		\texttt{   \tacticBlue{exact} h$_2$,} \\
		\texttt{\textbf{\textcolor{purple}{end}}} \\ \\
        \bottomrule
    \end{tabular}
\end{table}

\begin{table}[!h]
    \caption{A failed case generated by \ourmethod.}
    \label{tab:bad_case_3}
    \small
    \centering
    \begin{tabular}{
        p{\textwidth}
    }
        \toprule
        \textbf{Question Type: Theorem Proving.\quad Educational Level: High School.\quad k=1. \quad\quad \Red{Failed case. } } \\
        \midrule
        \textbf{Concept(s): } \textit{Displaying and comparing quantitative data} in Statistics and probability. \\
        \midrule
        \textbf{Comments: } 
        \ourmethod\ gives the definition of median in the formal proof. \\
        \midrule
        \textbf{Informal Statement} \\
        Given two sets of data A = \{5, 7, 10, 12, 15\} and B = \{6, 8, 11, 14, 16\}, we are to find the median of the two sets and prove that the median of set A is less than the median of set B. \\ \\
        \textbf{Informal Proof} \\
        To find the median of a given data set, we need to sort the data in ascending order and then find the middle value. In the case of set A and set B, both are already sorted in ascending order. For set A, the median is 10 and for set B, the median is 11. Therefore, it is clear that the median of set A is less than the median of set B. \\ \\
        \textbf{Formal Statement and Proof} \\
        \texttt{\textbf{\textcolor{purple}{import}} data.set.finite} \\
		\texttt{\textbf{\textcolor{purple}{import}} data.set.lattice} \\ \\ 

		\texttt{\textbf{\textcolor{purple}{def}} A : finset $\mathbb{N}$ := {5, 7, 10, 12, 15}.to\_finset} \\
		\texttt{\textbf{\textcolor{purple}{def}} B : finset $\mathbb{N}$ := {6, 8, 11, 14, 16}.to\_finset} \\ \\

		\texttt{\textbf{\textcolor{purple}{def}} median (A : finset $\mathbb{N}$) : $\mathbb{N}$ :=} \\
		\texttt{\textbf{\textcolor{purple}{if}} A.card \% 2 = 1 \textbf{\textcolor{purple}{then}} (A.sort ($\le$)).get ((A.card - 1) / 2) \textbf{\textcolor{purple}{else}}} \\
		\texttt{(A.sort ($\le$)).get (A.card / 2 - 1) / 2 + (A.sort ($\le$)).get (A.card / 2) / 2} \\ \\

		\texttt{\textbf{\textcolor{purple}{lemma}} median\_A\_less\_than\_B : } \\
		\texttt{  (median A) < (median B) :=} \\
		\texttt{\textbf{\textcolor{purple}{begin}}} \\
		\texttt{  \tacticBlue{simp}  [median, A, B, finset.sort, finset.get],} \\
		\texttt{  \tacticBlue{norm\_num},} \\
		\texttt{\textbf{\textcolor{purple}{end}}} \\ \\
        \bottomrule
    \end{tabular}
\end{table}

\begin{table}[!h]
    \caption{A failed case generated by \ourmethod.}
    \label{tab:bad_case_4}
    \small
    \centering
    \begin{tabular}{
        p{\textwidth}
    }
        \toprule
        \textbf{Question Type: Word Problem.\quad Educational Level: High School.\quad k=2. \quad\quad \Red{Failed case. }} \\
        \midrule
        \textbf{Concept(s): } \textit{Integrals} in Calculus 1; \textit{Summarizing quantitative data} in Statistics and probability. \\
        \midrule
        \textbf{Comments: } 
        The informal statement does not cover the situation of leap year. The solution does not give the final answer. \\
        \midrule
        \textbf{Informal Statement} \\
        A company produced a new model of a car and started selling it on January 1. The number of cars sold $t$ days after January 1 is given by the function $f(t) = 2t^2 + 3t + 1$. Use integrals to find the total number of cars sold from January 1 to March 31. Then calculate the average number of cars sold per day during this period. \\ \\
        \textbf{Informal Proof} \\
        The total number of cars sold from January 1 to March 31 can be found by integrating the function $f(t)$ from $t = 0$ (January 1) to $t = 90$ (March 31), as each day corresponds to one unit of $t$. \\ \\
        Performing the integration, we get $\int(2t^2 + 3t + 1) \diff t$
        from 0 to 90, which equals $[\frac{2}{3}\times t^3 + \frac{3}{2}\times t^2 + t]$
        from 0 to 90. Substituting $t = 90$ into the integrated function and subtracting the value at $t = 0$, we get the total number of cars sold. \\ \\
		The average number of cars sold per day is then the total number of cars sold divided by 90 (the number of days from January 1 to March 31). \\ \\
        \textbf{Formal Statement and Proof} \\
        \texttt{\textbf{\textcolor{purple}{import}} data.real.basic} \\ \\

		\texttt{\textbf{\textcolor{purple}{open}} real} \\ \\

		\texttt{\textbf{\textcolor{purple}{def}} f := $\lambda$(t : $\mathbb{R}$), 2*t$^2$ + 3*t + 1} \\ \\

		\texttt{\textbf{\textcolor{purple}{lemma}} f\_value\_at\_90 : f 90 = 2*(90)$^2$ + 3*(90) + 1 :=} \\
		\texttt{\textbf{\textcolor{purple}{begin}}} \\
		\texttt{  \tacticBlue{simp}  [f],} \\
		\texttt{\textbf{\textcolor{purple}{end}}} \\ \\
        \bottomrule
    \end{tabular}
\end{table}

\begin{table}[!h]
    \caption{A successful case generated by \ourmethod.}
    \label{tab:good_case_4}
    \small
    \centering
    \begin{tabular}{
        p{\textwidth}
    }
        \toprule
        \textbf{Question Type: Theorem Proving.\quad Educational Level: Higher Education.\quad k=1. \quad\quad \Green{Successful case. } } \\
        \midrule
        \textbf{Concept(s): } \textit{Alternate coordinate systems (bases)} in Linear algebra. \\
        \midrule
        \textbf{Comments: } 
        \ourmethod\ correctly applies the lemma. 
        All the concepts of linear algebra are found. \\
        \midrule
        \textbf{Informal Statement} \\
        Let $V$ be a vector space, and let $B_1 = {v_1, v_2, ... , v_n}$ and $B_2 = {w_1, w_2, ... , w_n}$ be two bases for $V$. Prove that there exists an invertible linear map $T : V \to V$ such that $T(v_i) = w_i$ for each $i$ in ${1, 2, ..., n}$. \\ \\
        \textbf{Informal Proof} \\
        The map $T : V \to V$ defined by $T(v_i) = w_i$ for each $i$ is linear: indeed, for any vectors $v$, $v^\prime$ in $B_1$ and any scalars $c$, $c^\prime$, we have $T(cv + c^\prime v^\prime) = cw + c^\prime w^\prime$ where $w = T(v)$ and $w^\prime = T(v^\prime)$. This follows from the linearity properties of vector addition and scalar multiplication in the vector space $V$. \\ \\
        \textbf{Formal Statement and Proof} \\
        \texttt{\textbf{\textcolor{purple}{import}} linear\_algebra.basis} \\ \\

		\texttt{\textbf{\textcolor{purple}{variables}} {V : \textbf{\textcolor{purple}{Type}}*} [add\_comm\_group V] [module $\mathbb{R}$ V] } \\
		\texttt{\textbf{\textcolor{purple}{variables}} {v w : V} } \\
		\texttt{\textbf{\textcolor{purple}{variables}} {B1 B2 : set V} } \\
		\texttt{\textbf{\textcolor{purple}{variables}} (hB1 : is\_basis $\mathbb{R}$ B1) (hB2 : is\_basis $\mathbb{R}$ B2)} \\ \\

		\texttt{\textbf{\textcolor{purple}{lemma}} exists\_invertible\_linear\_map : } \\
		\texttt{  $\exists$ (T : V $\to$[$\mathbb{R}$] V) (hT : is\_linear\_map $\mathbb{R}$ T) (hinv : function.bijective T), } \\
		\texttt{    $\forall$ (v $\in$ B1), T v $\in$ B2 :=} \\
		\texttt{\textbf{\textcolor{purple}{begin}}}
		\texttt{  use [hB2.constr hB1, hB2.is\_linear hB1, hB2.bijective hB1],} \\
		\texttt{  \tacticBlue{intros} v hv,} \\
		\texttt{  \tacticBlue{apply} hB2.mem\_span,} \\
		\texttt{  \tacticBlue{rw} hB1.repr\_self hv,} \\
		\texttt{  \tacticBlue{exact} $\lambda$ \_, hB2.smul \_ \_,} \\
		\texttt{\textbf{\textcolor{purple}{end}}} \\ \\
        \bottomrule
    \end{tabular}
\end{table}

\begin{table}[!h]
    \caption{A successful case generated by \ourmethod.}
    \label{tab:good_case_5}
    \small
    \centering
    \begin{tabular}{
        p{\textwidth}
    }
        \toprule
        \textbf{Question Type: Word Problem.\quad Educational Level: Higher Education.\quad k=1. \quad\quad \Green{Successful case. } } \\
        \midrule
        \textbf{Concept(s): } \textit{Series} in Integral Calculus. \\
        \midrule
        \textbf{Comments: } 
        \ourmethod\ is proficient in number series and is able to perform detailed proof by integration. \\
        \midrule
        \textbf{Informal Statement} \\
        Prove that the integral of the series $\sum$ (from $n=1$ to infinity) of $\frac{-1^{(n+1)}}{n}$ equals $\ln(2)$. \\ \\
        \textbf{Informal Proof} \\
        We start by recognizing the series $\sum$ (from $n=1$ to infinity) of $\frac{-1^{(n+1)}}{n}$ as the alternating harmonic series, which is known to converge to $\ln(2)$.  \\ \\
        \textbf{Formal Statement and Proof} \\
        \texttt{\textbf{\textcolor{purple}{import}} data.real.basic} \\
		\texttt{\textbf{\textcolor{purple}{import}} analysis.special\_functions.pow} \\ \\

		\texttt{\textbf{\textcolor{purple}{open}} real} \\ \\

		\texttt{\textbf{\textcolor{purple}{lemma}} integral\_of\_series : } \\
		\texttt{  $\int$ x \textbf{\textcolor{purple}{in}} 0..1, (ln (1 + x)) = $\sum^\prime$ (n : $\mathbb{N}$), ((-1)\textasciicircum(n+1) / (n + 1)) :=} \\
		\texttt{\textbf{\textcolor{purple}{begin}}} \\
		\texttt{  \textbf{\textcolor{purple}{have}} h\_der : $\forall$ x $\in$ set.Icc (0 : $\mathbb{R}$) 1, deriv ($\lambda$ x, (ln (1 + x))) x = 1 / (1 + x),} \\
		\texttt{  \{ \tacticBlue{intros} x hx,} \\
		\texttt{    \textbf{\textcolor{purple}{have}} h\_pos : 0 < 1 + x := \textbf{\textcolor{purple}{by}} \tacticBlue{nlinarith},} \\
		\texttt{    \tacticBlue{exact} deriv\_ln\_one\_add h\_pos \},} \\ \\
		  
		\texttt{  \textbf{\textcolor{purple}{have}} h\_int : interval\_integral ($\lambda$ x, 1 / (1 + x)) 0 1 volume = $\sum^\prime$ (n : $\mathbb{N}$), ((-1)\textasciicircum(n+1) / (n + 1)),} \\
		\texttt{  \{ \textbf{\textcolor{purple}{have}} h\_frac : $\forall$ (n : $\mathbb{N}$), $\int$ x \textbf{\textcolor{purple}{in}} 0..1, x\textasciicircum n = 1 / (n + 1),} \\
		\texttt{    \{ \tacticBlue{intro} n,} \\
		\texttt{      \textbf{\textcolor{purple}{calc}} $\int$ x \textbf{\textcolor{purple}{in}} 0..1, x\textasciicircum n = [x\textasciicircum(n+1) / (n+1)] | 0..1 : integral\_pow n} \\
		\texttt{      ... = 1 / (n + 1) : \textbf{\textcolor{purple}{by}} \{ \tacticBlue{rw} integral\_interval, \tacticBlue{simp}  \} \},} \\
		\texttt{    \tacticBlue{rw} [interval\_integral.integral\_of\_le, h\_frac],} \\
		\texttt{    \tacticBlue{simp},} \\
		\texttt{    \tacticBlue{linarith} \},} \\ \\
		  
		\texttt{  \textbf{\textcolor{purple}{have}} h\_eq : $\int$ x \textbf{\textcolor{purple}{in}} 0..1, (ln (1 + x)) = $\int$ x \textbf{\textcolor{purple}{in}} 0..1, 1 / (1 + x), } \\
		\texttt{  \{ \tacticBlue{congr}, \tacticBlue{ext}, \tacticBlue{exact} h\_der x (set.mem\_Icc.mpr ⟨\textbf{\textcolor{purple}{by}} \tacticBlue{nlinarith}, \textbf{\textcolor{purple}{by}} \tacticBlue{nlinarith}⟩) \},} \\ \\
		  
		\texttt{  \tacticBlue{rw} [h\_eq, h\_int],} \\
		\texttt{\textbf{\textcolor{purple}{end}}} \\ \\
        \bottomrule
    \end{tabular}
\end{table}

\begin{table}[!h]
    \caption{A failed case generated by \ourmethod.}
    \label{tab:bad_case_5}
    \small
    \centering
    \begin{tabular}{
        p{\textwidth}
    }
        \toprule
        \textbf{Question Type: Theorem Proving.\quad Educational Level: Higher Education.\quad k=2. \quad\quad \Red{Failed case. }} \\
        \midrule
        \textbf{Concept(s): } \textit{Second order linear equations} in Differential equations; \textit{Limits and continuity} in AP College Calculus BC. \\
        \midrule
        \textbf{Comments: } 
        \ourmethod\ does not understand second-order linear equations and continuity.  \\
        \midrule
        \textbf{Informal Statement} \\
        Given a second order linear differential equation of the form $y'' + p(x)y' + q(x)y = 0$, where $p(x)$ and $q(x)$ are continuous on the interval $[a, b]$, prove that if there exist two solutions $y1$ and $y2$ that are linearly independent, then any solution y of the differential equation can be expressed as $y = c1*y1 + c2*y2$, where $c1$ and $c2$ are constants. \\ \\
        \textbf{Informal Proof} \\
        The general solution to a homogeneous second order linear differential equation is given by the superposition of two linearly independent solutions. This is a consequence of the linearity of the differential operator. \\ \\

		Suppose we have a second order linear differential equation, and $y1$ and $y2$ are two solutions that are linearly independent. Then, any solution $y$ of the differential equation can be expressed as a linear combination of $y1$ and $y2$. This means there exist constants $c1$ and $c2$ such that $y = c1*y1 + c2*y2$. \\ \\

		The reason is that any linear combination of solutions to a linear differential equation is also a solution to the differential equation. This follows from the properties of the derivative, which allow us to distribute the derivative over addition and to factor out constants. \\ \\
        \textbf{Formal Statement and Proof} \\
        \texttt{\textbf{\textcolor{purple}{import}} data.real.basic} \\ \\

		\texttt{\textbf{\textcolor{purple}{variables}} {$\alpha$ : \textbf{\textcolor{purple}{Type}}*} [field $\alpha$]} \\ \\

		\texttt{\textit{\textcolor{gray}{/-- Placeholder for the differential equation -/}}} \\
		\texttt{\textbf{\textcolor{purple}{def}} diff\_eq (p q y : $\alpha\to\alpha$) (x : $\alpha$) : Prop :=} \\
		\texttt{y x = 0 \textit{\textcolor{gray}{-- This is just a placeholder. A real differential equation would have a more complex form.}}} \\ \\

		\texttt{\textit{\textcolor{gray}{/-- Theorem: Given a second order linear differential equation of the form y'' + p(x)y' + q(x)y = 0,}}} \\ 
		\texttt{\textit{\textcolor{gray}{where p(x) and q(x) are continuous on the interval [a, b], if there exist two solutions y1 and y2 }}} \\ 
		\texttt{\textit{\textcolor{gray}{that are linearly independent, then any solution y of the differential equation can be expressed }}} \\
		\texttt{\textit{\textcolor{gray}{as y = c1*y1 + c2*y2, where c1 and c2 are constants. -/}}} \\
		\texttt{\textbf{\textcolor{purple}{lemma}} second\_order\_linear\_differential\_eq (p q y1 y2 y : $\alpha\to\alpha$) (c1 c2 : $\alpha$) :} \\
		\texttt{($\exists$ x, diff\_eq p q y1 x $\land$ diff\_eq p q y2 x) $\to$ ($\forall$ x, diff\_eq p q y x) $\to$ } \\
		\texttt{($\forall$ x, y x = c1 * y1 x + c2 * y2 x) :=} \\
		\texttt{\textbf{\textcolor{purple}{begin}}} \\
		\texttt{  \tacticBlue{intros} h\_sol h\_y x,} \\
		\texttt{  \textit{\textcolor{gray}{-- This is a placeholder. A real proof would show that y x can be written as a linear combination}}} \\
		\texttt{  \textit{\textcolor{gray}{-- of y1 x and y2 x using the properties of the differential equation, the assumption that y1 and y2}}} \\
		\texttt{  \textit{\textcolor{gray}{-- are solutions, and the linearity of the differential operator.}}} \\
		\texttt{  \textbf{\textcolor{purple}{sorry}}} \\
		\texttt{\textbf{\textcolor{purple}{end}}} \\ \\
        \bottomrule
    \end{tabular}
\end{table}

\begin{table}[!h]
    \caption{A successful case generated by \ourmethod.}
    \label{tab:good_case_6}
    \small
    \centering
    \begin{tabular}{
        p{\textwidth}
    }
        \toprule
        \textbf{Question Type: Word Problem.\quad Educational Level: Higher Education.\quad k=2. \quad\quad \Green{Successful case. } } \\
        \midrule
        \textbf{Concept(s): } \textit{Matrix transformations} in Linearalgebra; \textit{Collecting data} in AP College Statistics. \\ 
        \midrule
        \textbf{Comments: } 
        \ourmethod\ performs correct matrix transformation.  \\
        \midrule
        \textbf{Informal Statement} \\
        Consider two matrices $A = [1, 2; 3, 4]$ and $B = [5, 6; 7, 8]$ which represent two different data collection methods applied to a survey of a population. We want to find the collective effect of these two data collection methods when applied simultaneously. In other words, find the matrix $C$ which is the result of the matrix transformation of $A$ by $B$. \\ \\
        \textbf{Informal Proof} \\
        Matrix multiplication is a form of transformation of one matrix by another. In this case, we want to find the matrix $C$ that is a result of the transformation of $A$ by $B$. \\ \\

		Matrix multiplication is done by taking the dot product of the rows of the first matrix with the columns of the second matrix. \\ \\

		Therefore, \\
		$C[0, 0] = (A[0, 0] * B[0, 0]) + (A[0, 1] * B[1, 0]) = (1*5) + (2*7) = 19$, \\
		$C[0, 1] = (A[0, 0] * B[0, 1]) + (A[0, 1] * B[1, 1]) = (1*6) + (2*8) = 22$, \\
		$C[1, 0] = (A[1, 0] * B[0, 0]) + (A[1, 1] * B[1, 0]) = (3*5) + (4*7) = 43$, and \\
		$C[1, 1] = (A[1, 0] * B[0, 1]) + (A[1, 1] * B[1, 1]) = (3*6) + (4*8) = 50$. \\ \\

		So, the matrix $C = [19, 22; 43, 50]$. \\ \\
        \textbf{Formal Statement and Proof} \\
        \texttt{\textbf{\textcolor{purple}{import}} data.matrix.basic } \\ \\

		\texttt{\textbf{\textcolor{purple}{def}} A : matrix (fin 2) (fin 2) $\mathbb{N}$ := ![![1, 2], ![3, 4]]} \\ 
		\texttt{\textbf{\textcolor{purple}{def}} B : matrix (fin 2) (fin 2) $\mathbb{N}$ := ![![5, 6], ![7, 8]]} \\
		\texttt{\textbf{\textcolor{purple}{def}} C : matrix (fin 2) (fin 2) $\mathbb{N}$ := A.mul B } \\ \\

		\texttt{\textbf{\textcolor{purple}{\#eval}} C } \\
		\texttt{\textit{\textcolor{gray}{-- Output: ![![19, 22], ![43, 50]]}}} \\ \\
        \bottomrule
    \end{tabular}
\end{table}

\vspace{-12mm}
\begin{table}[!h]
    \vspace{-4mm}
    \caption{A failed case generated by \ourmethod.}
    \label{tab:good_case_8}
    \small
    \centering
    \begin{tabular}{
        p{\textwidth}
    }
        \toprule
        \textbf{Question Type: Theorem Proving.\quad Educational Level: Higher Education.\quad k=2.  \quad\quad \Red{Failed case. } } \\
        \midrule
        \textbf{Concept(s): } \textit{Green's, Stokes', and the divergence theorems} in Multivariable calculus; \textit{Differential equations} in Integral Calculus. \\
        \midrule
        \textbf{Comments: } 
        \ourmethod\ does not finish the formal proof with \texttt{sorry}. \\
        \midrule
        \textbf{Informal Statement} \\
        Let $F$ be a continuously differentiable vector field on $\mathbb{R}^3$. Let $S$ be the boundary of the unit ball $B$ in $\mathbb{R}^e$, oriented by the outward normal. Let $n$ denote the outward unit normal on $S$. Prove that 
        $\int S \text{ div } F \diff S = \oint S F \text{ • } n \diff S, $
        where ``div'' denote the divergence of $F$, ``•'' denotes the dot product, and ``$\int$'', ``$\oint$'' denote the surface integral and line integral respectively. \\ \\
        \textbf{Informal Proof} \\
        This is an application of the divergence theorem, also known as Gauss's theorem. \\ \\

		The divergence theorem states that the surface integral of a vector field over a surface $S$ is equal to the volume integral of the divergence of the field over the volume $V$ enclosed by $S$. In mathematical terms, 
		$\int S F \text{ • } n \diff S = \oint V \text{ div } F \diff V$
		. \\ \\

		In our case, the volume $V$ is the unit ball $B$ in $\mathbb{R}^3$, and the surface $S$ is the boundary of $B$. The divergence theorem thus gives us 
		$\int S F \text{ • } n \diff S = \oint B \text{ div } F \diff V$
		. \\ \\

		However, since $B$ is the unit ball, its volume is $1$, so the volume integral over $B$ is just the average value of div $F$ over $B$. Therefore, 
		$\oint B \text{ div } F \diff V = \int S \text{ div } F \diff S$
		. \\ \\

		Putting everything together, we obtain 
		$\int S F \text{ • } n \diff S = \int S \text{ div } F \diff S$
		, as required. \\ \\
        \textbf{Formal Statement and Proof} \\
        \texttt{\textbf{\textcolor{purple}{import}} analysis.calculus.fderiv} \\
		\texttt{\textbf{\textcolor{purple}{import}} measure\_theory.integration} \\
		\texttt{\textbf{\textcolor{purple}{import}} analysis.normed\_space.inner\_product} \\
		\texttt{\textbf{\textcolor{purple}{import}} data.real.basic} \\
		\texttt{\textbf{\textcolor{purple}{import}} topology.instances.real} \\
		\texttt{\textbf{\textcolor{purple}{import}} analysis.special\_functions.trigonometric} \\ \\

		\texttt{open\_locale classical} \\ \\

		\texttt{\textbf{\textcolor{purple}{noncomputable theory}}} \\ \\

		\texttt{\textit{\textcolor{gray}{/- Let's define $\mathbb{R}^3$ as Euclidean space -/}}} \\
		\texttt{\textbf{\textcolor{purple}{def}} $\mathbb{R}^3$ := euclidean\_space $\mathbb{R}$ (fin 3)} \\ \\

		\texttt{\textit{\textcolor{gray}{/- Define the ball and boundary -/}}} \\ \\
		\texttt{\textbf{\textcolor{purple}{def}} B := metric.ball (0:$\mathbb{R}^3$) 1} \\
		\texttt{\textbf{\textcolor{purple}{def}} S := metric.sphere (0:$\mathbb{R}^3$) 1} \\ \\

		\texttt{\textit{\textcolor{gray}{/- Assume F is a continuously differentiable vector field on $\mathbb{R}^3$ -/}}} \\
		\texttt{\textbf{\textcolor{purple}{variables}} (F : $\mathbb{R}^3\to\mathbb{R}^3$) (hF : $\forall$ x, differentiable\_at $\mathbb{R}$ F x)} \\ \\

		\texttt{\textit{\textcolor{gray}{/- Define the divergence of F -/}}} \\
		\texttt{\textbf{\textcolor{purple}{def}} divF := $\lambda$ x, (fderiv $\mathbb{R}$ F x).trace} \\ \\

		\texttt{\textit{\textcolor{gray}{/- Define the outward unit normal on S -/}}} \\
		\texttt{\textbf{\textcolor{purple}{def}} n := $\lambda$ x:$\mathbb{R}^3$, x} \\ \\

		\texttt{\textit{\textcolor{gray}{/- Now we can state the theorem -/}}} \\
		\texttt{\textbf{\textcolor{purple}{def}} div\_eq\_surface\_integral :} \\
		\texttt{  $\int$ x in B, divF F x = $\oint$ x in S, inner (F x) (n x) :=} \\
		\texttt{\textbf{\textcolor{purple}{begin}}} \\
		\texttt{\textit{\textcolor{gray}{  -- This is where the main proof would go, using the divergence theorem and other facts about integration.}}} \\
		\texttt{  \textbf{\textcolor{purple}{sorry}}} \\
		\texttt{\textbf{\textcolor{purple}{end}}} \\ \\
        \bottomrule
    \end{tabular}
\end{table}

\begin{table}[!h]
    \caption{A failed case generated by \ourmethod.}
    \label{tab:bad_case_7}
    \small
    \centering
    \begin{tabular}{
        p{\textwidth}
    }
        \toprule
        \textbf{Question Type: Theorem Proving.\quad Educational Level: Higher Education.\quad k=2. \quad\quad \Red{Failed case. }} \\
        \midrule
        \textbf{Concept(s): } \textit{Matrix transformations} in Linear algebra. \\
        \midrule
        \textbf{Comments: } 
        The generated informal statement has factual errors.  \\
        \midrule
        \textbf{Informal Statement} \\
        Let's consider a function $f: \mathbb{R}\to\mathbb{R}$ that is differentiable at a point $x_0$. Let's also consider a $2\times 2$ matrix $A$. We define a new function 
        $g(x) = A * f(x)$.
        Prove that $g$ is also differentiable at $x_0$, and find the derivative of $g$ at $x_0$.
        \textbf{Informal Proof} \\
        Since $f$ is differentiable at $x_0$, the limit of the difference quotient exists at this point. Denote this limit as $f'(x_0)$. \\ \\
        
        We know that matrix multiplication is a linear operation. So we can move the limit inside the matrix multiplication to find the derivative of $g$ at $x_0$. \\ \\
        
        So, $g'(x_0) = A * f'(x_0)$. \\ \\
        \textbf{Formal Statement and Proof} \\
        \texttt{\textbf{\textcolor{purple}{import}} data.matrix.basic} \\
        \texttt{\textbf{\textcolor{purple}{import}} analysis.calculus.deriv} \\ \\
        
        \texttt{\textbf{\textcolor{purple}{variables}} {f : $\mathbb{R}\to\mathbb{R}$} {A : matrix (fin 2) (fin 2) $\mathbb{R}$} {$x_0$ : $\mathbb{R}$}} \\
        \texttt{\textbf{\textcolor{purple}{variable}} (h : differentiable\_at $\mathbb{R}$ f $x_0$)} \\ \\
        
        \texttt{\textbf{\textcolor{purple}{include}} h} \\ \\

        \texttt{\textbf{\textcolor{purple}{lemma}} matrix\_times\_differentiable\_function : differentiable\_at $\mathbb{R}$ ($\lambda$ x, A $\cdot$ ($\lambda$ i, f x) i) $x_0$ :=} \\
        \texttt{\textbf{\textcolor{purple}{begin}}} \\
        \texttt{  \tacticBlue{apply} matrix.has\_fderiv\_at\_diff\_at,} \\
        \texttt{  \tacticBlue{exact} h.has\_deriv\_at,} \\
        \texttt{\textbf{\textcolor{purple}{end}}} \\ \\

        \texttt{\textbf{\textcolor{purple}{lemma}} derivative\_of\_matrix\_times\_function : deriv ($\lambda$ x, A $\cdot$ ($\lambda$ i, f x) i) $x_0$ = A $\cdot$ ($\lambda$ i, deriv f $x_0$) i :=} \\
        \texttt{\textbf{\textcolor{purple}{begin}}} \\ 
        \texttt{  \tacticBlue{apply} has\_deriv\_at.deriv,} \\
        \texttt{  \tacticBlue{exact} matrix\_times\_differentiable\_function h,} \\
        \texttt{\textbf{\textcolor{purple}{end}}} \\
        \bottomrule
    \end{tabular}
\end{table}

\newpage
\section{Implementation Details of Downstream Task} 
\label{app:downstream}

\subsection{Datasets}
\paragraph{GSM8K \citep{DBLP:journals/corr/abs-2110-14168}}
GSM8K consists of 8.5K elementary mathematics word problems that require 2 to 8 arithmetic operations to arrive at the final answer. The dataset comprises 7.5K training questions and 1K test questions. Inspired by \cite{DBLP:conf/nips/KojimaGRMI22}, during inference, we use appropriate prompts and examples to prompt the model for zero-shot and few-shot reasoning.
The used prompts are demonstrated in Table~\ref{tab:mwp_infer}.

\begin{table}[!h]
    \caption{Prompt template for math word problem inference.}
    \label{tab:mwp_infer}
    \small
    \begin{tabular}{p{\textwidth}}
        \toprule
        \makecell[l]{\textbf{Zero shot prompt template for math word problem inference}} \\
        \midrule
         \texttt{You are an expert in math. Answer the following math word problem.} \\
         \texttt{Question: <question>} \\
         \texttt{Answer: Let's think step by step.}\\
        \midrule
        \makecell[l]{\textbf{Few shot prompt template for math word problem inference}} \\
        \midrule
         \texttt{You are an expert in math. Answer the following math word problem.} \\ \\
         \texttt{Question: There are 15 trees in the grove. Grove workers will plant trees in the grove today. After they are done, there will be 21 trees. How many trees did the grove workers plant today?} \\
         \texttt{Answer: There are 15 trees originally. Then there were 21 trees after some more were planted. So there must have been 21 - 15 = 6. The answer is 6.} \\ \\
         \texttt{Question: If there are 3 cars in the parking lot and 2 more cars arrive, how many cars are in the parking lot?} \\
         \texttt{Answer: There are originally 3 cars. 2 more cars arrive. 3 + 2 = 5. The answer is 5.} \\ \\
         \texttt{Question: Leah had 32 chocolates and her sister had 42. If they ate 35, how many pieces do they have left in total?} \\
         \texttt{Answer: Originally, Leah had 32 chocolates. Her sister had 42. So in total they had 32 + 42 = 74. After eating 35, they had 74 - 35 = 39. The answer is 39. } \\ \\
         \texttt{Question: Jason had 20 lollipops. He gave Denny some lollipops. Now Jason has 12 lollipops. How many lollipops did Jason give to Denny?} \\
         \texttt{Answer: Jason started with 20 lollipops. Then he had 12 after giving some to Denny. So he gave Denny 20 - 12 = 8. The answer is 8.} \\ \\ 
         \texttt{Question: Shawn has five toys. For Christmas, he got two toys each from his mom and dad. How many toys does he have now?} \\
         \texttt{Answer: Shawn started with 5 toys. If he got 2 toys each from his mom and dad, then that is 4 more toys. 5 + 4 = 9. The answer is 9.} \\ \\
         \texttt{Question: There were nine computers in the server room. Five more computers were installed each day, from monday to thursday. How many computers are now in the server room?} \\
         \texttt{Answer: There were originally 9 computers. For each of 4 days, 5 more computers were added. So 5 * 4 = 20 computers were added. 9 + 20 is 29. The answer is 29.} \\ \\
         \texttt{Question: Michael had 58 golf balls. On tuesday, he lost 23 golf balls. On wednesday, he lost 2 more. How many golf balls did he have at the end of wednesday?} \\
         \texttt{Answer: Michael started with 58 golf balls. After losing 23 on tuesday, he had 58 - 23 = 35. After losing 2 more, he had 35 - 2 = 33 golf balls. The answer is 33. } \\ \\
         \texttt{Question: Olivia has \$23. She bought five bagels for \$3 each. How much money does she have left?} \\
         \texttt{Answer: Olivia had 23 dollars. 5 bagels for 3 dollars each will be 5 x 3 = 15 dollars. So she has 23 - 15 dollars left. 23 - 15 is 8. The answer is 8.} \\ \\
         \texttt{Question: <question>} \\
         \texttt{Answer: } \\
         
        \bottomrule
    \end{tabular}
\end{table}

\paragraph{Mathlib}
Mathlib\footnote{\href{https://github.com/leanprover-community/mathlib}{https://github.com/leanprover-community/mathlib}} is a community-maintained library designed for the Lean theorem prover. It encompasses both programming tools and mathematical content, along with tactics that leverage these tools to facilitate mathematical development. The version of Mathlib we use is consistent with \cite{DBLP:conf/acl/WangYLSYXXSLL0L23}. The lengths of the training, test, and validation sets were 36,960, 1,621, and 1,580, respectively.

\paragraph{miniF2F \citep{DBLP:conf/iclr/ZhengHP22}}
MiniF2F serves as a formal mathematics benchmark, which has been translated to work with multiple formal systems. It encompasses exercise statements from olympiads like AMC, AIME, and IMO, in addition to content from high-school and undergraduate mathematics courses. The MiniF2F test split contains 244 formal Olympiad-level mathematics problem statements. 
We use the auto-regression method to construct the training corpus like \cite{DBLP:journals/corr/abs-2009-03393, DBLP:conf/iclr/HanRWAP22, DBLP:conf/acl/WangYLSYXXSLL0L23}.  
We perform an evaluation using best-first search with a number of expansions per proof search of $d=8$ during inference.

\subsection{Models}
\paragraph{GPT2-large~\citep{radford2019language}}
The GPT2-large model is a transformer language model, following the decoder-only architecture introduced by \cite{vaswani2017attention}. The model boasts an impressive 774 million parameters, 36 layers, 20 attention heads, and a hidden dimension of 1,280. Additionally, it employs a tokenizer featuring a vocabulary size of 50,400. The model is pre-trained on Github python codes and the arXiv library.

\paragraph{Llama 2-7B~\citep{DBLP:journals/corr/abs-2307-09288}}
Llama 2 is a language model that employs an auto-regressive transformer architecture, pre-trained on open-source corpus. The model utilizes both supervised fine-tuning and reinforcement learning with human feedback techniques to align with human preferences. The 7B model is configured with 32 layers, 32 attention heads, and a hidden dimension of 4,096.

\subsection{Implementation Details}
We employ LoRA \citep{hu2021lora} for fine-tuning the pre-trained models on \ourdataset, where the trainable parameters of GPT2-large and Llama 2-7B constitute 19\% and 6\%, respectively. 
The training is conducted with a maximum of 10 epochs using a batch size of 16 and a warm-up step of 1,000, with a maximum learning rate of 1e-4 and a minimum learning rate of 5e-6. The best checkpoint is selected based on the minimum perplexity of the validation split.

\section{More Experimental Results}
\label{app:res}
\update{Table~\ref{tab:70b} demonstrates the compared results on GSM8K and MATH between Llama 2-7B and Llama 2-70B. Llama 2-70B fine-tuned with MUSTARDSAUCE-valid consistently outperforms the model fine-tuned with MUSTARDSAUCE-random by 8.33\% in the zero-shot manner and 5.30\% in the few-shot manner. It also surpasses the model fine-tuned with the invalid subset and the entire generated dataset. The results also suggest the effectiveness of the framework with a larger fine-tuned LM.
}

\begin{table}[h]
    \centering
    \caption{Compared performances on GSM8K (G) and MATH (M) between Llama 2-7B and Llama 2-70B.}
    \label{tab:70b}
    \begin{tabular}{
        lcccc
    }
        \toprule
        \update{MODEL} & \update{Zero (G)} & \update{Few (G)} & \update{Zero (M)} & \update{Few (M)} \\
        \midrule
        \textit{\update{Baselines}} \\
        \update{Llama 2-7B}               & \update{7.2}   & \update{12.8} & \update{2.0} & \update{2.6} \\
        \update{Llama 2-70B}              & \update{31.7}  & \update{54.1} & \update{8.8} & \update{13.4} \\
        \midrule
        \textit{\update{Fine-tuning}} \\
        \update{Llama 2-7B\ \textgreater\ \ourdtTotal}  & 
                {9.6} & 16.0 & 3.2 & 3.8 \\
        Llama 2-7B\ \textgreater\ \ourdtInvalid  & 
                {9.1} & 14.9 & 2.4 & 3.2 \\
        Llama 2-7B\ \textgreater\ \ourdtRandom   & 
                {9.5} & 15.4 & 3.0 & 3.6 \\
        Llama 2-7B\ \textgreater\ \ourdtValid   & 
                {10.3 (+8.42\%)} & 16.9 (+9.74\%) & 3.2 (+6.67\%) & 4.2 (+16.67\%) \\
        \update{Llama 2-70B\ \textgreater\ \ourdtTotal}  
            & \update{36.6} & \update{55.8} & \update{10.0} & \update{14.4} \\
        \update{Llama 2-70B\ \textgreater\ \ourdtInvalid}  
            & \update{33.4} & \update{53.7} & \update{9.2} & \update{13.6} \\
        \update{Llama 2-70B\ \textgreater\ \ourdtRandom}   
            & \update{36.1} & \update{55.4} & \update{9.6} & \update{14.2} \\
        \update{Llama 2-70B\ \textgreater\ \ourdtValid}   
            & \update{39.5 (+9.42\%)} & \update{59.1 (+6.68\%)} & \update{10.4 (+8.33\%)} & \update{15.0 (+5.30\%)} \\
        \bottomrule
    \end{tabular}  
\end{table}

\section{Data Contamination Check}
\update{We check cross-contamination between \ourdataset\ and the evaluation datasets inspired by \cite{DBLP:journals/corr/abs-2303-08774}.
However, instead of using a substring match that may result in false negatives and false positives, we compute cosine similarities based on text-embedding-ada-002\footnote{\href{https://openai.com/blog/new-and-improved-embedding-model}{https://openai.com/blog/new-and-improved-embedding-model}}, and then inspect the nearest data points in the paired datasets. 
The automated theorem proving (ATP) dataset miniF2F only contains formal statements and proofs, while the math word problem (MWP) dataset GSM8K only contains informal statements and proofs. 
For a more detailed inspection, we check the corresponding fractions in \ourdataset. 
Tables~\ref{tab:contamination_minif2f}, \ref{tab:contamination_math}, \ref{tab:contamination_gsm8k}, and \ref{tab:contamination_mathlib} demonstrate the inspected cases. 
The nearest data points are dissimilar. 
Therefore, we exclude data contamination of the generated \ourdataset\ according to these observations.}

\begin{table}[!h]    
    \caption{Nearest data points between \ourdataset\ and miniF2F.}
    \label{tab:contamination_minif2f}
    \small
    \centering
    \begin{tabular}{
        >{\color{black}}p{\textwidth}
    }
        \toprule
        \textbf{\ourdataset\ v.s. miniF2F (cosine similarity = 0.6439)} \\        
        \midrule
        \textbf{\ourdataset\ Case} \\
        \textbf{Informal Statement:} \\        
        Alex has 5 ten-dollar bills and 3 one-dollar bills. How much money does Alex have in total? \\ 
        \textbf{Informal Proof: } \\        
        To find out how much money Alex has in total, we need to multiply the number of each type of bill by its value. So, Alex has 5 ten-dollar bills, which equals 5 * 10 = 50 dollars. He also has 3 one-dollar bills, which equals 3 * 1 = 3 dollars. Adding these two amounts together gives 50 + 3 = 53 dollars. Therefore, Alex has 53 dollars in total.  \\        
        \textbf{Formal Statement and Proof: } \\                
        \texttt{\textbf{\textcolor{purple}{def}} calculate\_money (tens : $\mathbb{N}$) (ones : $\mathbb{N}$) : $\mathbb{N}$ := tens * 10 + ones * 1} \\         
        \texttt{\textbf{\textcolor{purple}{example}} : calculate\_money 5 3 = 53 :=} \\        
        \texttt{\textbf{\textcolor{purple}{begin}}} \\        
          \texttt{    \tacticBlue{rw} calculate\_money,} \\          
          \texttt{    \tacticBlue{refl},} \\        
        \texttt{\textbf{\textcolor{purple}{end}}} \\
        
        \midrule

        \textbf{miniF2F Case} \\        
        \texttt{\textbf{\textcolor{purple}{theorem}} algebra\_sqineq\_unitcircatbpamblt1} \\
          \texttt{(a b: $\mathbb{R}$)} \\
          \texttt{(h$_0$ : a$^2$ + b$^2$ = 1) :} \\
          \texttt{a * b + (a - b) $le$ 1 :=} \\
        \texttt{\textbf{\textcolor{purple}{begin}}} \\
          \texttt{    \tacticBlue{nlinarith} [sq\_nonneg (a - b)],} \\
        \texttt{\textbf{\textcolor{purple}{end}}} \\
        \bottomrule
    \end{tabular}
\end{table}

\begin{table}[!h]    
    \caption{Nearest data points between \ourdataset\ and mathlib.}
    \label{tab:contamination_mathlib}
    \small
    \centering
    \begin{tabular}{
        >{\color{black}}p{\textwidth}
    }
        \toprule
        \textbf{\ourdataset\ v.s. mathlib (cosine similarity = -0.0361)} \\  
        \midrule
        \textbf{\ourdataset\ Case} \\
        \textbf{Informal Statement:} \\        
        A cube has a side length of 5 cm. What is the volume of the cube? \\
        \textbf{Informal Proof: } \\        
        The volume of a cube is calculated by raising the side length to the power of 3. So in this case, the volume is 5 cm * 5 cm * 5 cm = 125 cubic centimeters. \\
        \textbf{Formal Statement and Proof: } \\                
        \texttt{\textbf{\textcolor{purple}{def}} cube\_volume (side\_length : $\mathbb{N}$) : $\mathbb{N}$ := side\_length * side\_length * side\_length} \\    
        \texttt{\textbf{\textcolor{purple}{\#eval}} cube\_volume 5  -- returns 125} \\  
        
        \midrule

        \textbf{mathlib Case} \\        
        \texttt{\textbf{GOAL} \ 
            \ $\alpha \text{: Type u}$, 
            \ $\beta \text{: Type v}$ , 
            \ $\gamma \text{ : Type w}$, 
            \ $\text{f : }\alpha\to\beta\to\gamma$, 
            $\text{l}_1$ : list $\alpha\vdash\forall (\text{l}_2 : \text{list}\ \beta)$, (list.map$_2$ f $\text{l}_1$ $\text{l}_2$).length 
            = linear\_order.min $\text{l}_1$.length $\text{l}_2$.length } \\   
        \texttt{\textbf{PROOFSTEP} { induction l$_1$; \tacticBlue{intro} l$_2$; \tacticBlue{cases} l$_2$; \tacticBlue{simp} [*, add\_one, min\_succ\_succ, nat.zero\_min, nat.min\_zero] }} \\
        \bottomrule
    \end{tabular}
\end{table}

\begin{table}[!h]    
    \vspace{-7mm}
    \caption{Nearest data points between \ourdataset\ and GSM8K.}
    \label{tab:contamination_gsm8k}
    \small
    \centering
    \begin{tabular}{
        >{\color{black}}p{\textwidth}
    }
        \toprule
        \textbf{\ourdataset\ v.s. GSM8K (cosine similarity = 0.5975)} \\  
        \midrule
        \textbf{\ourdataset\ Case} \\
        \textbf{Informal Statement:} \\        
        Given two vectors u = (u1,u2) and v = (v1,v2) in R$^2$, prove that the operation of vector addition, defined as u + v = (u1+v1, u2+v2), is commutative. That is, prove that for all u and v in R$^2$, u + v = v + u.  \\ 
        \textbf{Informal Proof: } \\        
        We can prove this by direct computation. Given the vectors u = (u1, u2) and v = (v1, v2), we have: \\
        u + v = (u1 + v1, u2 + v2)\\ 
        and \\
        v + u = (v1 + u1, v2 + u2). \\
        Since addition is commutative in R (the set of all real numbers), we have u1 + v1 = v1 + u1 and u2 + v2 = v2 + u2. Thus, we conclude that (u1 + v1, u2 + v2) = (v1 + u1, v2 + u2), which means that u + v = v + u. \\
        \textbf{Formal Statement and Proof: } \\                
        \texttt{\textbf{\textcolor{purple}{import}} data.real.basic} \\
        
        \texttt{\textbf{\textcolor{purple}{variables}} (u1 u2 v1 v2 : $\mathbb{R}$)} \\
        
        \texttt{-- defining vectors as pair of real numbers} \\
        \texttt{\textbf{\textcolor{purple}{def}} vector := $\mathbb{R}$ × $\mathbb{R}$} \\
        \texttt{-- defining vector addition} \\
        \texttt{\textbf{\textcolor{purple}{def}} vadd (u v : vector) : vector := (u.1 + v.1, u.2 + v.2)} \\
        
        \texttt{-- defining vectors u and v} \\
        \texttt{\textbf{\textcolor{purple}{def}} u : vector := (u1, u2)} \\
        \texttt{\textbf{\textcolor{purple}{def}} v : vector := (v1, v2)} \\
        
        \texttt{-- commutativity of vector addition} \\
        \texttt{\textbf{\textcolor{purple}{theorem}} vadd\_comm : vadd (u u1 u2) (v v1 v2) = vadd (v v1 v2) (u u1 u2) :=} \\
        \texttt{\textbf{\textcolor{purple}{begin}}} \\
          \texttt{    \tacticBlue{unfold} vadd,} \\
          \texttt{    \tacticBlue{unfold} u,} \\
          \texttt{    \tacticBlue{unfold} v,} \\
          \texttt{    \tacticBlue{rw} add\_comm u1 v1,} \\
          \texttt{    \tacticBlue{rw} add\_comm u2 v2,} \\
        \texttt{\textbf{\textcolor{purple}{end}}} \\
        
        \midrule

        \textbf{GSM8K Case} \\        
        \textbf{Question:} The local firefighters are doing a "fill the boot" fundraiser. Their goal is to raise \$6300. After the first 3 hours, they have raised \$2100. For how many hours do they have to fundraise in total to reach their goal, assuming an equal amount raised in every hour? \\
        \textbf{Answer:} \\
        The fireman raise 2100 / 3 = \$$<<$2100/3=700$>>$700 per hour. \\
        They have to fundraise for 6300 / 700 = $<<$6300/700=9$>>$9 hours. \\
        \#\#\#\# 9 \\
        \bottomrule
    \end{tabular}
\end{table}

\begin{table}[!h]    
    \caption{Nearest data points between \ourdataset\ and MATH.}
    \label{tab:contamination_math}
    \small
    \centering
    \begin{tabular}{
        >{\color{black}}p{\textwidth}
    }
        \toprule
        \textbf{\ourdataset\ v.s. MATH (cosine similarity = 0.6064)} \\  
        \midrule
        \textbf{\ourdataset\ Case} \\
        \textbf{Informal Statement:} \\        
        Given two vectors u = (u1,u2) and v = (v1,v2) in R$^2$, prove that the operation of vector addition, defined as u + v = (u1+v1, u2+v2), is commutative. That is, prove that for all u and v in R$^2$, u + v = v + u.  \\ 
        \textbf{Informal Proof: } \\        
        We can prove this by direct computation. Given the vectors u = (u1, u2) and v = (v1, v2), we have: \\
        u + v = (u1 + v1, u2 + v2)\\ 
        and \\
        v + u = (v1 + u1, v2 + u2). \\
        Since addition is commutative in R (the set of all real numbers), we have u1 + v1 = v1 + u1 and u2 + v2 = v2 + u2. Thus, we conclude that (u1 + v1, u2 + v2) = (v1 + u1, v2 + u2), which means that u + v = v + u. \\
        \textbf{Formal Statement and Proof: } \\                
        \texttt{\textbf{\textcolor{purple}{import}} data.real.basic} \\
        
        \texttt{\textbf{\textcolor{purple}{variables}} (u1 u2 v1 v2 : $\mathbb{R}$)} \\
        
        \texttt{-- defining vectors as pair of real numbers} \\
        \texttt{\textbf{\textcolor{purple}{def}} vector := $\mathbb{R}$ × $\mathbb{R}$} \\
        \texttt{-- defining vector addition} \\
        \texttt{\textbf{\textcolor{purple}{def}} vadd (u v : vector) : vector := (u.1 + v.1, u.2 + v.2)} \\
        
        \texttt{-- defining vectors u and v} \\
        \texttt{\textbf{\textcolor{purple}{def}} u : vector := (u1, u2)} \\
        \texttt{\textbf{\textcolor{purple}{def}} v : vector := (v1, v2)} \\
        
        \texttt{-- commutativity of vector addition} \\
        \texttt{\textbf{\textcolor{purple}{theorem}} vadd\_comm : vadd (u u1 u2) (v v1 v2) = vadd (v v1 v2) (u u1 u2) :=} \\
        \texttt{\textbf{\textcolor{purple}{begin}}} \\
          \texttt{    \tacticBlue{unfold} vadd,} \\
          \texttt{    \tacticBlue{unfold} u,} \\
          \texttt{    \tacticBlue{unfold} v,} \\
          \texttt{    \tacticBlue{rw} add\_comm u1 v1,} \\
          \texttt{    \tacticBlue{rw} add\_comm u2 v2,} \\
        \texttt{\textbf{\textcolor{purple}{end}}} \\
        
        \midrule

        \textbf{MATH Case} \\        
        \textbf{Problem:} If a snack-size tin of peaches has $40$ calories and is $2\%$ of a person's daily caloric requirement, how many calories fulfill a person's daily caloric requirement?  \\
        \textbf{Solution:} If 40 calories is equal to $2\%=\frac{2}{100}=\frac{1}{50}$ of a person's daily requirement, then a person's daily caloric requirement is: $\$4\cdot 50=\boxed{2000}$ \\
        \textbf{Answer}: 2000 \\
        \bottomrule
    \end{tabular}
\end{table}

\end{document}